%% file: main.tex
\documentclass{article} 

\usepackage[preprint]{icml2026}

\input{math_commands.tex}

\usepackage{hyperref}
\usepackage{url}
\usepackage{graphicx}
\usepackage{subfigure}
\usepackage{algorithm}
\usepackage{algpseudocode}
\usepackage{amsthm}
\usepackage{booktabs}

\usepackage{subcaption}
\usepackage{array}

\usepackage[export]{adjustbox}

\newtheorem{defn}{Definition}
\newtheorem{thm}{Theorem}

\icmltitlerunning{Matrix-Driven Identification and Reconstruction of LLM Weight Homology}

\begin{document}


\twocolumn[
  \icmltitle{Matrix-Driven Identification and Reconstruction of LLM Weight Homology}



  \icmlsetsymbol{equal}{*}

  \begin{icmlauthorlist}
    \icmlauthor{Ruichong Zhang}{tsinghua_univ}
    \icmlauthor{Daniel Goldstein}{eleuther,featherless}
  \end{icmlauthorlist}

  \icmlaffiliation{tsinghua_univ}{Qiuzhen College, Tsinghua University, Beijing, China}
  \icmlaffiliation{eleuther}{EleutherAI}
  \icmlaffiliation{featherless}{Recursal AI}
  


  \icmlkeywords{Machine Learning, ICML}

  \vskip 0.3in
]
\printAffiliationsAndNotice{}

\begin{abstract}
We propose Matrix-Driven Identification and Reconstruction (MDIR), a SOTA large language model homology method that accurately detects weight correspondences between models and provides rigorous $p$-value estimation of the statistical significance of these correspondences. Our method does not require model inference, and allows the detection of unattributed reuse or replication of model weights even on low-resource devices as it compares only a single pair of matrices at a time. We leverage matrix analysis, polar decomposition, and Large Deviation Theory (LDT) to achieve accurate reconstruction of weight relationships between models. Notably, MDIR is the first method to achieve perfect scores on both Area-Under-Curve (AUC) and accuracy metrics across different source models on LeaFBench~\citep{shao2025soklargelanguagemodel}.

\end{abstract}

\section{Introduction}

\begin{table}[h]
\centering
\caption{Performance comparison of LLM homology methods via LeaFBench~\citep{shao2025soklargelanguagemodel} across different source models. The baseline methods are: Gradient~\citep{wu2025gradientbasedmodelfingerprintingllm},  REEF~\citep{zhang2025reef}, HuRef~\citep{zeng2024huref}, and PDF~\citep{yoon2025intrinsicfingerprintllmscontinue}. Best results are bolded.}
\begin{adjustbox}{max width=1.0\columnwidth}
\begin{tabular}{lccccc}
\hline
Method & ACC & AUC  & pAUC & MD  & TPR@ \\
 &   &  &   &  & 1\%FPR \\
\hline
Gradient & 0.812 & 0.798 & 0.708 & 0.672 & 0.374 \\
REEF & 0.892 & 0.896 & 0.832 & 2.091 & 0.634 \\
HuRef & 0.992 & 0.994 & 0.971 & \textbf{3.154} & 0.943 \\
PDF & 0.996 & 0.995 & 0.996 & 1.741 & 0.992 \\
\textbf{MDIR} & \textbf{1.000} & \textbf{1.000}  & \textbf{1.000} & 2.834 & \textbf{1.000} \\
\hline
\end{tabular}
\end{adjustbox}

\end{table}

\begin{table}[h]
\centering
\caption{Ability of different LLM homology methods to compare different model dimensionalities, different tokenizers, to output layer-wise relationships, to output patterns for visual inspection, to run without requiring full model inference, and to output true statistical $p$-values instead of unadjusted similarity measures. }
\begin{adjustbox}{max width=1.0\columnwidth}
\begin{tabular}{lcccccc}
\hline
Method & Dim. & Tokenizer & Layer & Visual & NoInf & $p$\\
\hline
Gradient & x & - & x & x & x & x \\
REEF & \checkmark & \checkmark & - & \checkmark & x & x\\
HuRef & x & \checkmark & x & x & \checkmark & x\\
PDF & \checkmark & \checkmark & x & x & \checkmark & x\\
SeedPrints & \checkmark & x & x & x & x & \checkmark \\
\textbf{MDIR} & \textbf{\checkmark} & \textbf{\checkmark} & \textbf{\checkmark} & \textbf{\checkmark} & \textbf{\checkmark} & \textbf{\checkmark} \\
\hline
\end{tabular}
\end{adjustbox}

\end{table}

Advances in large language models (LLMs) have led to widespread development and adaptation of models trained on massive datasets.
Recently, concerns have grown about the unattributed reuse or replication of model weights,
especially in cases involving direct copying, up-cycling \citep{yao2024masked,he2025upcyclinglargelanguagemodels}, pruning \citep{ma2023llmpruner,meta2024llama32}, or continual pretraining.
The scale and complexity of LLMs make detection of model weight homology particularly challenging.

\subsection{Problem Statement}
In this paper, we focus solely on determining the weight homology between pairs of large language models.
Specifically, given two LLMs $A$ and $B$, with their parameters denoted as $\theta_A$ and $\theta_B$,
we aim to determine whether $A$ and $B$ exhibit a relationship in their weights, based solely on the statistical properties of $\theta_A$ and $\theta_B$. 
This task can be formulated as a binary classification problem $\Psi$, where the inputs are $(\theta_A, \theta_B)$ and the output is $\Psi(\theta_A, \theta_B) \in \{0, 1\}$ Here, $1$ indicates that the two models are homologous in weights, while $0$ indicates unrelated. 
The weight relationships we consider include, but are not limited to, the following cases:
\begin{itemize}
    \item \textbf{Fine-tuning}: Various kinds of SFT and RL included;
    \item \textbf{Continual Pretraining}: Training the model with more data in the general domain, sometimes as much as trillions of tokens;
    \item \textbf{Upcycling} \citep{yao2024masked,he2025upcyclinglargelanguagemodels}: Continual pretraining with a larger model, especially an MoE model, with weights initialized from a smaller base model (usually dense);
    \item \textbf{Pruning} \citep{ma2023llmpruner,meta2024llama32}: Removing certain channels or neurons of a base model to obtain a smaller model;
    \item \textbf{Transformation}: Should include permutations and even general orthogonal/unitary transformations;
    \item \textbf{Comprehensive}: Combination of all these above.
\end{itemize}
It is worth noting that, in our setting, two models pretrained with same initialization (e.g., same random seed) are considered homologous, while two differently initialized models trained on same data, no matter how much, are still not considered homologous. The vast number of parameters (on the order of billions or even trillions) as problem input and the scarcity of examples makes weight homology a challenging problem.

\subsection{Taxonomy}
Current methods for detecting weight homology can be roughly classified into two categories: black-box approaches that do not require access to model weights (models are only available via API service), and white-box approaches with direct access to weights. With weight access and enabling analysis of weights and activations, white-box methods can further be classified into two subcategories: those requiring full model forwarding and those operating solely on weights.

\textbf{Black-box Methods.} Black-box methods \citep{xu-etal-2024-instructional,pasquini2025llmmapfingerprintinglargelanguage} usually rely on specific prompts or synthetic text fed into the model. For instance, during pretraining, models are trained on certain question-answer pairs $\{(q_i, a_i)\}$ to maximize $p_\theta(a_i \mid q_i)$. Downstream models exhibiting anomalously high $p_{\theta'}(a_i \mid q_i)$ (or low perplexity) relative to a naive baseline may then be flagged as derived from the original. 
This approach proved useful in a reported case involving Llama3-V and MiniCPM-o v2.6 \citep{OpenBMB_MiniCPM_o_Issue196}, 
where rare oracle bone inscriptions were used as questions and their corresponding modern Chinese characters as answers. 
However, the effectiveness of this method depends heavily on specific prompts or data tested, which may create an inherent bias; some prompts might be proprietary and not known to external users. Moreover, it has been shown recently that many models of different origin exhibit a surprising level of homogeneity, producing nearly identical outputs under many open-ended prompts \citep{jiang2025artificial}. Since model homogeneity (similarity in model output or policy) can occur without homology (similarity in weights stemming from evolution from the same weights), this finding may limit the practical applicability of black-box methods for weight homology detection.

\textbf{White-box Methods.} White-box methods, such as REEF \citep{zhang2025reef}, HuRef \citep{zeng2024huref}, 
Intrinsic Fingerprinting (\citet{he2025upcyclinglargelanguagemodels}, also known as PDF) and concurrent work SeedPrints \citep{tong2025seedprintsfingerprintstellseed} determine model weight similarity via examining either weights directly, or activations under certain inputs. Similarity can be determined either via comparing the weights/activations directly, or by reducing those into a fingerprint first and then comparing the fingerprints. An ideal model similarity measure should be invariant to common transformations (e.g., permutation and scaling) and remain stable even after extensive continual pretraining. However, most of these methods provide a similarity metric rather than binary output, and converting to binary output requires calibration of threshold on the AUC-ROC curve. 
Moreover, while these methods effectively reveal similarity through ``fingerprints,'' they generally lack the ability to reconstruct the weight correspondence mapping (e.g., transformations such as neuron permutation or channel scaling) between models.

\subsection{Our Method} 
We introduce Matrix-Driven Identification-Reconstruction (MDIR), a novel white-box method that operates exclusively on model weights without requiring inference. Instead of focusing on similarity scores or fingerprints, MDIR directly computes $p$-values to determine the statistical significance of weight homology. Leveraging Singular Value Decomposition (SVD), polar decomposition, and Large Deviation Theory (LDT) \citep{DemboZeitouni1998}, our method provides rigorous statistical $p$-values and reconstructs weight correspondence mappings when homology is identified. MDIR achieves state-of-the-art performance in detecting weight homology, achieving perfect 100\% scores on both AUC and accuracy in LeaFBench.
The overall process of MDIR is listed as follows:
\begin{enumerate}
    \item Obtain the orthogonal relation matrix between a pair of weights, firstly the embeddings;
    \item Apply Hungarian algorithm to identify any permutations present;
    \item Leverage LDT to obtain statistical $p$-value that this relationship occurred randomly;
    \item If $p$-value is indeed significant, repeat steps 1, 2, 3 on remaining weight matrices.
\end{enumerate}

We find that in cases where an adversary has not taken purposeful obfuscation measures, analysis of the embeddings alone typically reveals positive homology results with extremely high statistical significance.

\begin{algorithm}[htbp]
    \caption{MDIR on embeddings $E, E'$}
    \label{alg:embed}
    \begin{algorithmic}[1]
    \State \textbf{def} $\mathrm{Ortho}(A) \gets A(\sqrt{A^\text{T} A})^{-1}$
    \State \textbf{def} $\mathrm{PValue}(P, U, d) \gets d! \cdot \exp(-(\Tr(P U^{\text{T}}))^2 / 2)$
    \State $\mathcal{C} \gets $ Set of common vocabulary tokens
    \State $\tilde{U} \gets \mathrm{Ortho}(E[\mathcal{C}, :]^{\text{T}} E'[\mathcal{C}, :])$
    \State $P_E \gets \arg\max_{P \in \mathrm{Perm}(\mathrm{D_{emb}})} \Tr(P \tilde{U}^{\text{T}})$
    \State $U_E \gets P$ if $p_E < p_0$ else $\tilde{U}$
    \State $p_E \gets \mathrm{PValue}(P_E, \tilde{U}, \mathrm{D_{emb}})$
    \end{algorithmic}
\end{algorithm}

\section{Motivation}

\subsection{Invariant Transformations Preserve Functionality}

Modern LLMs are massively overparameterized, meaning many distinct parameter configurations $\theta$ and $\theta'$ can produce identical input-output behavior: $f_\theta = f_{\theta'}$ even when $\theta \neq \theta'$. These function-preserving configurations form what we call the \textbf{invariant space} of $\theta$:
\[
\mathcal{M}_{\text{inv}}(\theta) = \{ \theta' \in \mathcal{M}_{\text{arch}} \mid f_{\theta'} = f_{\theta} \},
\]
where $\mathcal{M}_{\text{arch}}$ is the parameter space of the model architecture. 
However, the structure of the space $\mathcal{M}_{\text{inv}}(\theta)$ is often complicated. It may even vary in dimension (for example, its dimension can be higher when there are $n$ neurons sharing the same input weights in $\theta$).

We seek for a set of transformations, such as orthogonal transformations on the backbone and attention heads, that form a continuous group $G$ that acts on the weights without changing the model's behavior (i.e., the orbit is always generating a submanifold of $\mathcal{M}_{\text{inv}}(\theta)$). We call such transformations \textbf{totally invariant}:
\begin{defn}
We call a Lie group $G$ acting on $\mathcal{M}_{\text{arch}}$ \textit{totally invariant} if:
\begin{enumerate}
    \item Every $g \in G$ preserves the input-output function: $f_{g \theta} = f_\theta$;
    \item $G$ acts isometrically (preserves distances in parameter space).
\end{enumerate}
\end{defn}

The probability that the weights could have randomly ended up near a location on this group can be determined with extremely high confidence if the group is of high dimension. This is because Large Deviation Theory implies probability dominated by an exponential quadratic term related to the trace of a matrix with this dimension. We construct such a high dimensional group $G$ explicitly for Transformer architectures in Section \ref{invar_gqa}.

\textbf{Key Insight: Training Trajectories Preserve $G$-Coordinates. }
Under idealized conditions (infinite numerical precision and $G$-invariant optimizers), training dynamics are confined to the \textit{quotient space} $\mathcal{M}_{\text{arch}} / G$. That is, the component of the parameters along the $G$-orbit (i.e., within the invariant group) remains unchanged throughout training.

To see why, consider a local orthogonal reparameterization near $\theta$: $(\alpha, \beta)$, where $\alpha$ parametrizes directions within $G\theta$ (the orbit), and $\beta$ parametrizes orthogonal directions that actually affect the function. Since the loss is invariant to $\alpha$, its gradient along $\alpha$ is zero:
\[
\frac{\partial \text{Loss}}{\partial \alpha^{(i)}} = 0 \quad \forall i.
\]
Thus, $\alpha$ remains constant at its initialization value.

If $G$ is an orthogonal subgroup, any orthogonal invariant optimizer is also $G$-invariant. 
This accounts for classical SGD, its momentum variants, Muon \citep{jordan2024muon}, Adam-mini \citep{zhang2025adamminiusefewerlearning}, and others. The Adam(W) optimizer is harder to analyze and is not fully orthogonal invariant due to element-wise second moment. However, since optimizers are designed to find an optimal solution that minimizes loss by following the gradient, we believe that it is likely that the optimization trajectory largely follows the gradient directions, respecting the fact that the gradient along $\alpha$ is zero. 
We find that in practice our method produces exceptionally strong results even for models trained for trillions of additional tokens using AdamW.

\textbf{Implication for Homology Detection.}
Suppose two models $\theta_1, \theta_2$ are derived from the same initialization (e.g., one is fine-tuned from the other). Then their $G$-components $g_1, g_2$ should satisfy $g_1^{-1} g_2 \approx e$, where $e$ is the identity element of the group $G$. 
Intuitively, this means that the transformation aligning $\theta_1$ to $\theta_2$ lies close to the identity transformation.
In contrast, independently initialized models will have $g_1^{-1} g_2$ distributed randomly across $G$. This suggests a simple homology detection criterion:
\[
\text{If } d_G(g_1^{-1} g_2, e) \text{ is small, } \theta_1 \text{ and } \theta_2 \text{ are likely homologous,}
\]
where $d_G(\_, e)$ can simply be measured by Frobenius inner product, which is equivalent to measuring the trace ($\langle g, e\rangle_F = \Tr(g)$, since $\langle g-e, g-e\rangle_F = \langle g, g\rangle_F + \langle e, e\rangle_F -2\Tr(g) = 2(n-\Tr(g))$). 

This principle directly enlightens our method: Using matrix SVD and polar decomposition techniques, we compute $g_1^{-1} g_2$ and measure its deviation from the identity via the trace.
Subsequently, this deviation can be converted to statistical significance of $p$-value via Large Deviation Theory.

\section{Methodologies}
\subsection{An Invariant Transformation Group for GQA}
\label{invar_gqa}
We assume that model $A$ is the original model and model $B$ the adversary, their parameters denoted as $\theta_A$ and $\theta_B$ respectively. 
Both models adopt a decoder-only Transformer architecture \citep{vaswani2017attention,radford2019language},  
with word embeddings and unembeddings, Grouped Query Attention (GQA) \citep{ainslie2023gqatraininggeneralizedmultiquery},  
MLP layers with up and down projections \citep{shazeer2020gluvariantsimprovetransformer}, and RMSNorm layers \citep{zhang2019rootmeansquarelayer}.  

We select GQA for our analysis framework, as GQA represents the most prevalent form of attention mechanism  
in modern Transformers. Both Multi-Head Attention (with an expansion rate of $1$) \citep{vaswani2017attention}  
and Multi-Query Attention (with one key-value head per layer) \citep{shazeer2019fasttransformerdecodingwritehead}  
can be treated as special cases of GQA.

We enumerate below a set of possible output-invariant orthogonal transformations on the main matrices of a GQA Transformer. This set has high dimension and can therefore be analyzed with high statistical certainty. In practice, it is often acceptable to analyze only a subset of these transformations. The embeddings are of particular importance, especially when an adversarial opponent intent on obfuscation is not present. 

Our method is able to identify many sub-combinations from of this set, but not all. For example, although we can correctly identify scaling combined with permutation, other combinations might be able to elude detection with our current numerical methods. We leave exploration of this to future work.

Note that our enumeration does not include all possible transformations, the set of which is astronomically large and may include uncountably many discontinuous transformations.  In particular, it does not cover any non-universally invariant transformations that cause invariant outputs for some specific choices of weights but not others.

\textbf{Transformations in the Attention Module.} For simplicity, we only consider one layer of attention, namely layer $\ell$. Assume that Model $B$ is equivalent to $A$ under certain transformations:
\[
\begin{aligned}
\theta_{A,\ell} &= \{ Q, K, V, O \} \\
\theta_{B,\ell} &= \{ Q', K', V', O' \}.
\end{aligned}
\]

The linear weight transformation from $\theta_A$ to $\theta_B$ may take the following form:
\[
\begin{aligned}
Q' &= U_Q Q W_Q, \quad K' = U_K K W_K, \\
V' &= U_V V W_V, \quad O' = W_O^{-1} O U_O^{-1},
\end{aligned}
\]
where $U_Q, U_K, U_V, U_O$ and $W_Q, W_K, W_V, W_O$ are transformation matrices applied to the original weights. These matrices represent modifications introduced during model adaptation.  

We refer to $U_Q, U_K, U_V, U_O$ as \textit{outer transformations}, which typically correspond to operations such as rotations, permutations, or scaled orthogonal transformations.  
Since both $U_Q, U_K, U_V$ and $Q, K, V$ operate on normalized vectors:
\[
\mathrm{RMSNorm}(x') U_\square = \mathrm{RMSNorm}(x), \quad \square \in \{ Q, K, V \},
\]
this implies that $U_Q = U_K = U_V$ (for simplicity, the equality holds up to a scalar multiple. Since scalars commute with all matrices, we move the scalar into the discussion of the inner transformations), and they are all orthogonal matrices. In the context of Lie groups, we denote $U = U_Q = U_K = U_V \in \mathrm{O}(\mathrm{EmbDim})$. 

Let $y$ be the original output of the attention module, and $y' = y U_O^{-1}$ be the new output. Because the input of the next feedforward network or attention module must match, we have
\[
\begin{aligned}
\mathrm{RMSNorm}(x'+y U_O^{-1}) U 
&= \mathrm{RMSNorm}(x'U+y U_O^{-1}U) \\
&= \mathrm{RMSNorm}(x+y) \ (\forall \ x, y)
\end{aligned}
\]
This implies $U_O^{-1}U = I$ and $U = U_Q = U_K = U_V =U_O$.

\textbf{Inner transformations.} For the \textit{inner transformations} $W_Q, W_K, W_V, W_O$, the situation becomes more complex due to the presence of attention heads and nonlinear transformations (e.g., Softmax) across channels.  
Not all orthogonal transformations are permissible.

\textbf{A General Form.}
Comprehensively, the \textit{outer transformations} are directly attached to the model backbone channels (the place where the residual is added), and should be identical across all layers. 
The \textit{inner transformations}, on the other hand, can be different for each layer.
Let $E \in \mathbb{R}^{\mathrm{VocabSize} \times \mathrm{EmbDim}}$ denote the vocabulary embedding matrix (and $F$ the unembedding matrix) of model $A$,  
and $E' \in \mathbb{R}^{\mathrm{VocabSize} \times \mathrm{EmbDim}}$ the vocabulary embedding for model $B$ (and $F'$ the unembedding).
\[
\begin{aligned}
    &E' = E U^{\text{T}}, \quad &&F' = U F, \\
    &Q'_\ell = U Q_\ell W_{Q, \ell}, \quad &&K'_\ell = U K_\ell W_{K, \ell}, \\
    &V'_\ell = U V_\ell W_{V, \ell}, \quad &&O'_\ell = W^{-1}_{O, \ell} O_\ell U^{\text{T}},\\
\end{aligned}
\]
where $W_{\square, \ell}, \ \square \in \{Q, K, V, O\}$ are the inner transformations with respect to layer $\ell$. In almost all real-world scenarios (including all models in LeaFBench), these transformations are all identity. However, while extremely rare in practice, inner transformations may not be identity. There can also be permutations, reflections and other transformations between heads both within a GQA group and across groups. We describe the details of this in Appendix~\ref{sec:inner_transformations_attention}.

\subsection{Solving the Transformations}
In an idealized setting, we have $E' = E U^{\text{T}}$ for the vocabulary embeddings of models $A$ and $B$. 
However, in practical scenarios, the adversary might have further trained its model, perturbing the weight of $E'$.
Following the established conventions, we have:
\[
E' = E U^{\text{T}} + N_E,
\]
where $U \in \mathbb{R}^{\mathrm{EmbDim} \times \mathrm{EmbDim}}$ is an orthogonal matrix,  
and $N_E$ represents the additional perturbation introduced by training or noise injection.

To minimize the difference between $E'$ and $E X^{\text{T}}$, we solve the following optimization problem:
\[
\begin{aligned}
&\quad\min_{X \in \mathrm{O}(\mathrm{EmbDim})} \lVert E' - E X^{\text{T}} \rVert_F^2 \\
&= \min_{X \in \mathrm{O}(\mathrm{EmbDim})} \left\langle E' - E X^{\text{T}}, E' - E X^{\text{T}} \right\rangle_F.
\end{aligned}
\]

Expanding the Frobenius norm yields:
\[
\begin{aligned}
    &\quad\arg\min_{X \in \mathrm{O}(\mathrm{EmbDim})} \lVert E' - E X^{\text{T}} \rVert_F^2 \\
    &= \arg\min_{X \in \mathrm{O}(\mathrm{EmbDim})} \left( \lVert E' \rVert_F^2 + \lVert E \rVert_F^2 - 2 \left\langle E X^{\text{T}}, E' \right\rangle_F \right) \\
    &= \arg\max_{X \in \mathrm{O}(\mathrm{EmbDim})} \left\langle E X^{\text{T}}, E' \right\rangle_F \\
    &= \arg\max_{X \in \mathrm{O}(\mathrm{EmbDim})} \Tr \left(E X^{\text{T}} E'^{\text{T}} \right) \\
    &= \arg\max_{X \in \mathrm{O}(\mathrm{EmbDim})} \Tr \left((E'^{\text{T}} E) X^{\text{T}} \right).
\end{aligned}
\]

We denote $\tilde{U}$ as the solution to this optimization problem. 
From the trace maximization property, the solution of $\tilde{U}$ is equal to the orthogonal factor in the polar decomposition of $(E'^{\text{T}} E)$.  
Note that $\tilde{U}$ is not the ground truth of $U$, but rather a close approximation. To reconstruct the actual $U$, we seek for special structural patterns lying behind $\tilde{U}$.
For example, if $\tilde{U}$ is sufficiently close to a permutation matrix $P \in \mathrm{Perm}(\mathrm{EmbDim})$, identifiable via:
\begin{equation}
\begin{aligned}
    \label{eqmainformula}
    P &= \argmax_{P \in \mathrm{Perm}(\mathrm{EmbDim})} \Tr(P \tilde{U}^{\text{T}}) \\
    &= \argmax_{P \in \mathrm{Perm}(\mathrm{EmbDim})} \Tr\left(P \ \mathrm{Ortho} (E'^{\text{T}} E)^{\text{T}}\right),
\end{aligned}
\end{equation}
then we may safely assert that $P$ is almost equal to the ground truth of $U$.
This problem is equivalent to solving maximum bipartite matching or the linear sum assignment problem \citep{scipy_linear_sum_assignment},  
which can be computed in up to $O(n^3)$ time using the Hungarian algorithm. 
To reduce computational cost, one may also use a faster heuristic: compute the positions of the maximal element in each row and verify whether they form a valid permutation.

A unique determination of $\tilde{U}$ requires $(E'^{\text{T}} E)$ to be non-degenerate and full-rank, which necessitates the vocabulary size to satisfy $\mathrm{VocabSize} \geq \mathrm{EmbDim}$.  
This condition is easily satisfied, as LLMs typically have much larger vocabularies than embedding size.

It is worth noting that,
our formula \ref{eqmainformula} is already immune to simple attacks, including scaling combined with permutation. See Appendix~\ref{sec:preservation_under_scaling_and_permutation} for details. 

\textbf{Changed Tokenizer.}
When model $B$ uses a different tokenizer, the embedding matrices $E$ and $E'$ are defined over different vocabularies. 
However, a substantial set of tokens $\mathcal{C}$, including ASCII bytes, common subwords (e.g., \texttt{is}, \texttt{take}), and morphemes (e.g., \texttt{-tion}), is typically shared. 
In contextualized representations, the meaning (and thus the embedding vector) of a token is determined by its usage across billions of contexts~\citep{mikolov2013distributed}. 
Consequently, even after independent training, the embeddings of shared tokens in homologous models remain approximately aligned up to the global transformation $U$. 
Let $\mathcal{C}$ denote the set of all common tokens. We estimate $\tilde{U}$ as:
\[
\tilde{U} = \arg\max_{X \in \mathrm{O}(\mathrm{EmbDim})} \Tr \left( \left(E'[\mathcal{C}, :]^{\text{T}} E[\mathcal{C}, :]\right) X^{\text{T}} \right).
\]
Thus, $\tilde{U}$ corresponds to the orthogonal part in the polar decomposition of $(E'[\mathcal{C}, :]^{\text{T}} E[\mathcal{C}, :])$.

\subsection{Estimating $p$-value}
After identifying a permutation matrix $P$ as $P = \arg\max_P\Tr(P \tilde{U}^{\text{T}})$,
we need a statistical criterion to determine whether our identification is significant.  
While visual inspection of $P$ and $\tilde{U}$ can provide qualitative evidence,  
the value of $\Tr(P \tilde{U}^{\text{T}})$ itself serves as a strong quantitative indicator.

\textbf{Null Hypothesis.} Our null hypothesis assumes that models $A$ and $B$ are not homologous,  
and there is no apparent similarity between their weights. Specifically, we assume that $\tilde{U}$ is uniformly distributed over the orthogonal group $\mathrm{O}(n)$ according to the Haar measure.  
This assumption is justified because under the null hypothesis, there is no systematic relationship between the weights of $A$ and $B$, and any observed alignment would be purely coincidental (See Appendix~\ref{app:uniform} for a justification).

Below, we take $n=\mathrm{EmbDim}$ for short. Under the null hypothesis, the probability measure $\mathrm{d}\mathbb{P}$ should be uniform across all admissible transformations,
and the distribution of $\tilde{U}$ should be uniform over $\mathrm{O}(n)$.

Now, fix $P_0$ as an arbitrary permutation matrix. We estimate the probability of $\Tr(P_0 \tilde{U}^{\text{T}}) \geq c$, denoted as:
\[
f(c) := \mathbb{P}\left[\Tr(P_0 \tilde{U}^{\text{T}}) \geq c\right] = \mathbb{P}\left[\Tr(\tilde{U}) \geq c\right],
\] 
since $P_0 \tilde{U}^{\text{T}}$ and $\tilde{U}$ are both uniformly distributed. 

With $n!$ possible permutations, only the one maximizing $\Tr(P \tilde{U}^{\text{T}})$ is chosen. To account for $n!$ comparisons across all permutations, we apply the conservative Bonferroni correction \citep{bonferroni_mathworld}, which adjusts the $p$-value by multiplying by $n!$. This yields:
\[
p \le n! \cdot \mathbb{P}\left[ \Tr(P_0 \tilde{U}^{\text{T}}) \ge c \right] = n! \cdot f(c).
\]
We estimate the $p$-value based on the evaluation of $f(c)$. Large Deviation Theory implies $f(c) \le K\exp(-c^2/2)$ for some constant $K$. Thus, an upper bound is established as
$ \log p \le \log (n!) - c^2 / 2 + \epsilon $. In practice, we set $\epsilon = 0$ for simplicity.

When $A$ and $B$ are homologous, $\Tr(P_0 \tilde{U}^{\text{T}})$ scales linearly with $n$ (e.g., $c \approx 0.4n$), so the quadratic term $-c^2/2$ dominates $\log(n!)$, resulting in a highly significant $p$-value.
Please refer to Appendix~\ref{appendix_estimate_p} for a detailed derivation.

\section{Experiments}
\subsection{Results on LeaFBench}
We evaluate MDIR on LeaFBench \citep{shao2025soklargelanguagemodel} and compare across four existing white-box methods: Gradient~\citep{wu2025gradientbasedmodelfingerprintingllm},  REEF~\citep{zhang2025reef}, HuRef~\citep{zeng2024huref}, PDF~\citep{yoon2025intrinsicfingerprintllmscontinue} and report the overall metrics. Since our method does not involve similarity as a primary metric, we report $-\log_{10}(p)$ as our ``similarity measure'' for threshold calibration, where $p$ is the $p$-value. MDIR is the first method to achieve perfect scores on both Area-Under-Curve (AUC) and accuracy.

While MDIR does not report similarity measure, obscuring the meaning of Mahalanobis distance (MD, \citet{DEMAESSCHALCK20001}), the perfect score of Area-Under-Curve suggests a clear separation threshold between true positives and false negatives. It is to be expected, as our $p$-values are true statistical measurements, that a threshold of even $0.001$ is already of very high significance. Moreover, upon close inspection, we found that setting $p$ threshold from $0.001$ to $10^{-125,105}$ does not affect final classification results. This implies any reasonable threshold (e.g., $10^{-10}$) can be set, and will hardly affect the final decision. 

\subsection{More Comparisons}
Since LeaFBench lacks certain model combinations, such as pruning, upscaling, extremely small and large models, and non-transformers, we further select 25 representative models for our comparison to evaluate the effectiveness of our MDIR method. For each pair of models, we compute the trace as $\max \Tr(P \tilde{U}^{\text{T}})$ via $P=\texttt{linear\_sum\_assignment}(\tilde{U})$ \citep{scipy_linear_sum_assignment}.
\begin{figure}[htbp]
    \centering
        \includegraphics[width=1.03\linewidth]{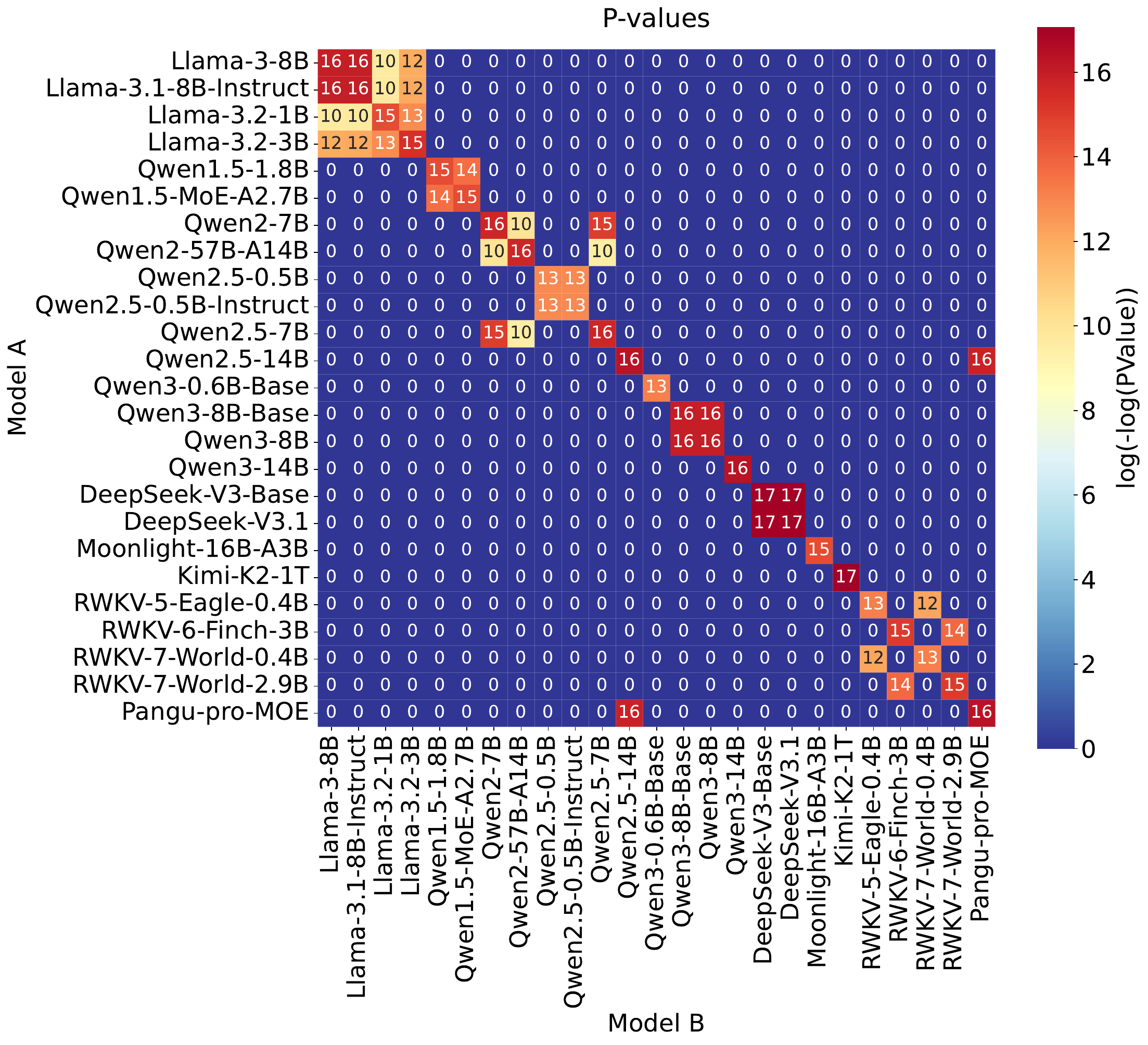}
    \caption{Comparison. We use $\log(-\log(p))$ to crop the $p$ values for better visualization. Value $0$ indicates no observable significance.}
    \label{fig:overall}
\end{figure}
From Figure \ref{fig:overall}, we observe that MDIR for homology detection is self-consistent — it fulfills the requirements of an equivalence relation: reflexivity, symmetry and transitivity.
It has also correctly identified all known homology relations, including the following types:
\begin{itemize}
    \item Instruction fine tuning and continual pretraining: Qwen2.5-0.5B-Instruct \citep{qwen2025qwen25technicalreport}, Llama-3.1-8B-Instruct \citep{meta_llama3_1_2024} and DeepSeek-V3.1 \citep{deepseekai2025deepseekv3technicalreport} are known to have trained from their predecessors;
    \item Pruning: Llama-3.2-1B and Llama-3.2-3B are pruned from Llama-3.1-8B \citep{meta2024llama32};
    \item Upcycling: Qwen1.5-MoE-A2.7B \citep{qwen_moe} and Qwen2-57B-A14B \citep{yang2024qwen2technicalreport} are upcycled from Qwen1.5-1.8B and Qwen2-7B, respectively;
    \item Non-transformer models: RWKV-7-World-0.4B and 2.9B \citep{peng2025rwkv7gooseexpressivedynamic} are upgraded from RWKV-5-Eagle-0.4B and RWKV-6-Finch-3B \citep{peng2024eagle} respectively;
    \item Independently developed models (models that are designed and trained from scratch): Moonlight-16B-A3B \citep{liu2025muonscalablellmtraining} and Kimi-K2 \citep{kimiteam2025kimik2openagentic}.
    \item Models of unclear origin: Pangu-Pro-MOE \citep{pangu_pro_moe_2025}; 
    \item Unclear phylogeny: It is unclear from the technical report whether Qwen2.5-7B is derived from Qwen2-7B by continual training.
\end{itemize}

It is worth mentioning that our significance threshold is set purely \textit{a priori}, based on the theoretical $p$-value bound, without any post-hoc calibration on known positive/negative pairs.

\subsection{Layer-level Relationship}
MDIR detects homology at both model and layer levels by reconstructing layer correspondence between a base model and its upscaled or pruned versions.

\begin{figure}[htbp]
    \centering
    
    \subfigure[MDIR]{
        \includegraphics[width=0.82\linewidth]{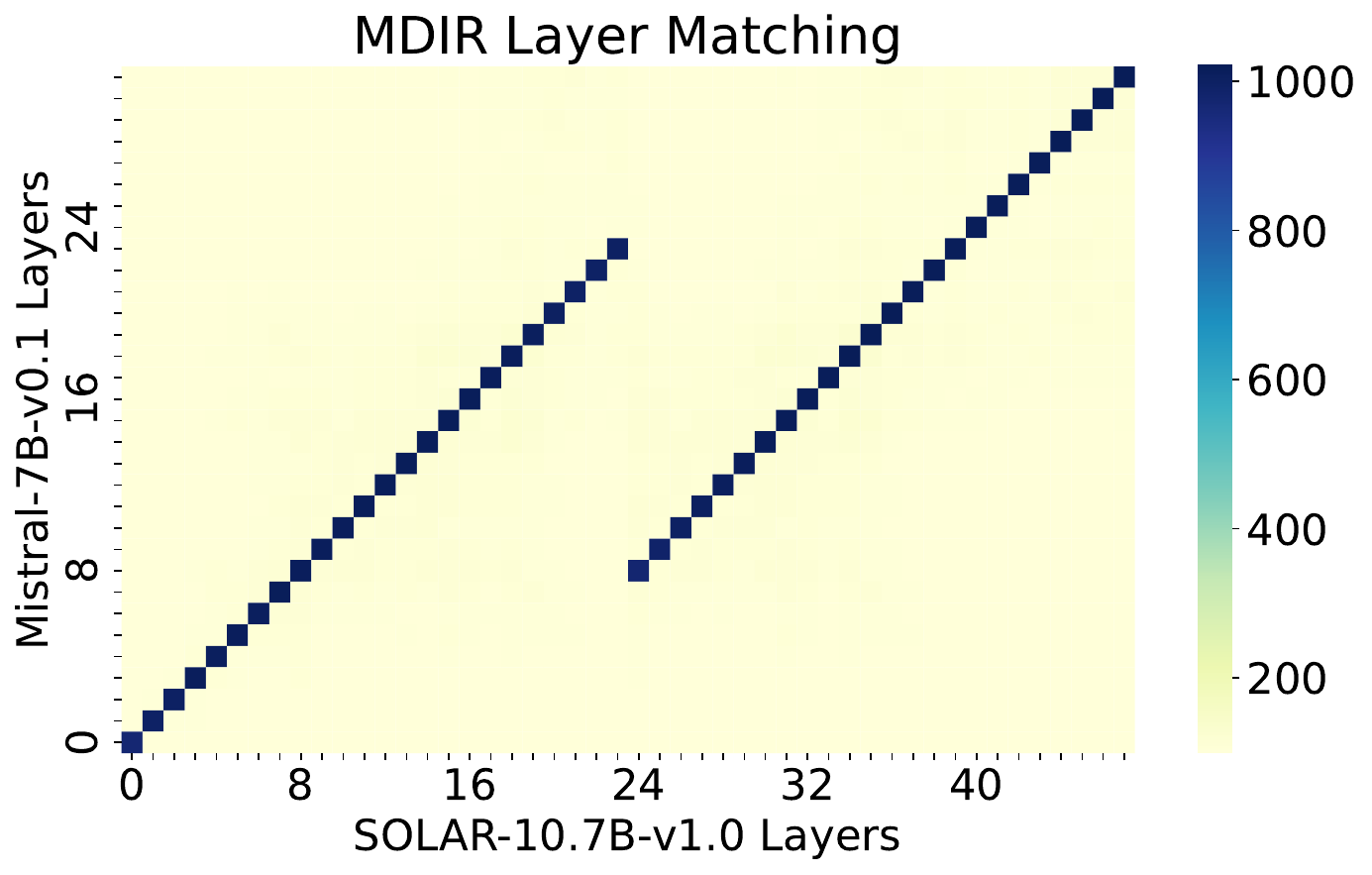}
        \vspace{-1em}
    }
    \vspace{-0.2em}
    \subfigure[REEF]{
        \includegraphics[width=0.78\linewidth]{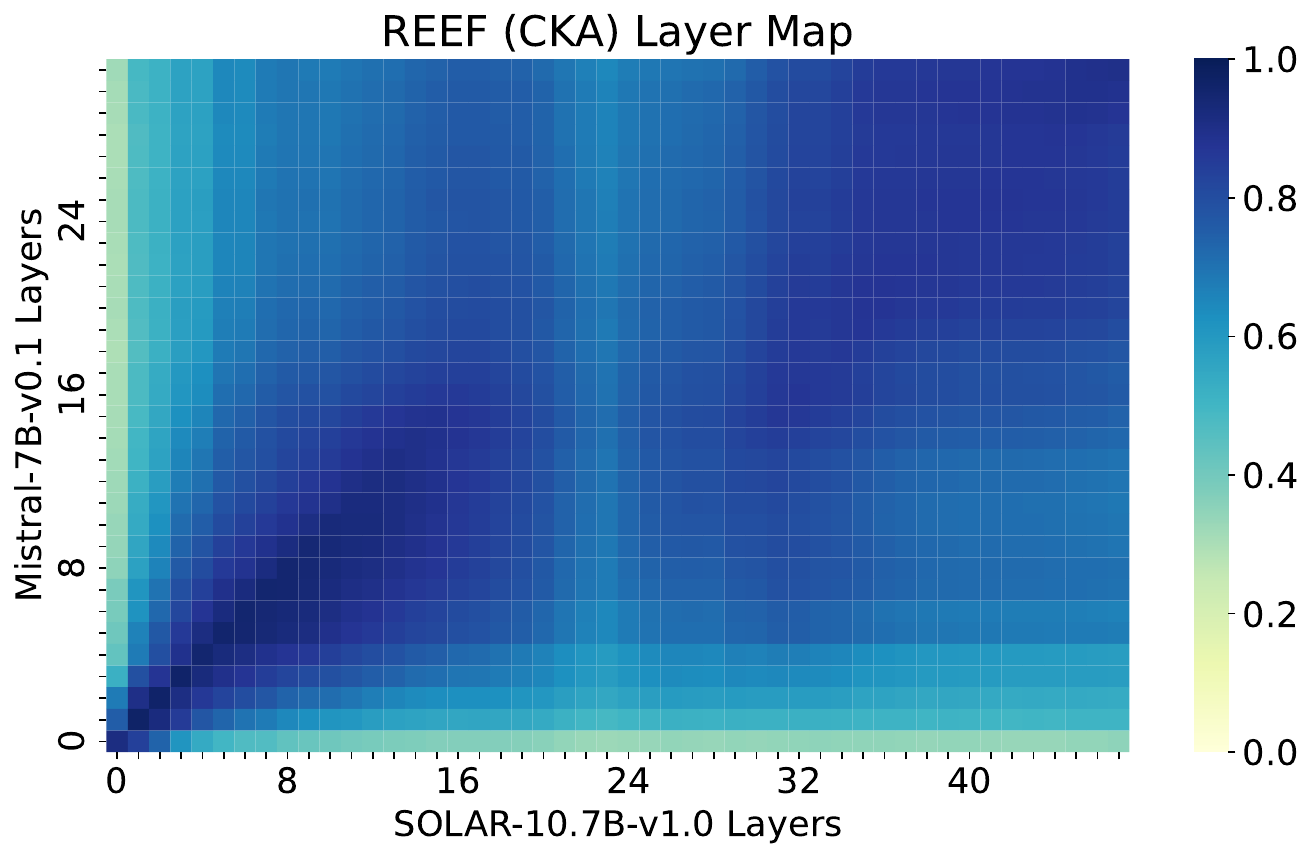}
    }
    \caption{Layerwise mapping of MDIR vs. REEF on SOLAR-10.7B and Mistral-7B-v0.1.}
    \vspace{-1em}
    \label{fig:solar}
\end{figure}

\textbf{Identifying a layer-upscaled model.}
We use SOLAR-10.7B \citep{kim2024solar107bscalinglarge}, depth-scaled from Mistral-7B-v0.1 \citep{jiang2023mistral7b}, as the test case. We select attention $V$ matrix as the representative for layer-level homology comparison, and compare our method against REEF. Results in Figure~\ref{fig:solar} confirm MDIR's strong, unambiguous signals, accurately identifying that SOLAR-10.7B takes the first 24 then last 24 layers of Mistral-7B-v0.1, consistent with its technical report. In contrast, REEF only produces a blurry, less reliable pattern.

\textbf{Identifying layer-pruned model.} See Appendix~\ref{sec_comparison}.

\subsection{Robustness}
We test our MDIR method against two deliberate adversarial attacks: by injecting noise, and by changing the tokenizer and retraining vocabulary embeddings.

\textbf{Noise Injection.} We test the robustness of MDIR by injecting increasing levels of Gaussian noise into the embedding matrices of two models: Qwen3-0.6B-Base and Llama-3.2-3B. The noise is scaled as $\text{NoiseLevel} 
\cdot \mathrm{RMS}(W)$ where $W$ is the original matrix, to ensure fair comparison across scales. We evaluate our method’s trace, normalized trace (trace divided by the model dimension), two baseline similarity measures: HuRef's cosine similarity and REEF’s Central Kernel Alignment \citep{kornblith2019similarityneuralnetworkrepresentations}, 
as well as model functionality, measured via LAMBADA perplexity \citep{paperno-etal-2016-lambada} and MMLU accuracy \citep{hendrycks2021measuringmassivemultitasklanguage}.
Results are summarized in Table~\ref{tab:noising}.

\begin{table}[h]
\centering
\caption{Metrics along with Noise Injection for Two Models}
\label{tab:noising}
\begin{subtable}
\centering
\small
(a) Qwen3-0.6B-Base
\begin{adjustbox}{max width=\linewidth}
\begin{tabular}{lccccc}
\toprule
Noise Level & 0 & 0.1 & 0.3 & 1.0 & 3.0 \\
\midrule
Trace (MDIR) & 1024 & 1023.98 & 1023.82 & 1021.96 & 1005.6 \\
Normalized Trace & 1 & 0.99998 & 0.9998 & 0.998 & 0.982 \\
Cosine Sim. & 1 & 0.995 & 0.958 & 0.707 & 0.316 \\
CKA (REEF) & 1 & 0.997 & 0.973 & 0.582 & 0.291 \\
LAMBADA PPL & 9.6 & 11.0 & 24.7 & 31{,}412 & $4\times10^{16}$ \\
MMLU Acc (\%) & 50.4 & 49.7 & 40.0 & 27.1 & 25.7 \\
\bottomrule
\end{tabular}
\end{adjustbox}
\label{subtab:qwen}
\end{subtable}

\vspace{1.2em}

\begin{subtable}
\centering
\small
(b) Llama-3.2-3B
\begin{adjustbox}{max width=\linewidth}
\begin{tabular}{lccccc}
\toprule
Noise Level & 0 & 0.1 & 0.3 & 1.0 & 3.0 \\
\midrule
Trace (MDIR) & 3072 & 3071.77 & 3069.88 & 3048.44 & 2859.64 \\
Normalized Trace & 1 & 0.99992 & 0.9993 & 0.992 & 0.931 \\
Cosine Sim. & 1 & 0.995 & 0.958 & 0.707 & 0.316 \\
CKA (REEF) & 1 & 0.9986 & 0.985 & 0.721 & 0.455 \\
LAMBADA PPL & 3.95 & 3.98 & 4.36 & 37.9 & $5\times10^{12}$ \\
MMLU Acc (\%) & 54.1 & 53.7 & 51.3 & 26.0 & 24.9 \\
\bottomrule
\end{tabular}
\end{adjustbox}
\label{subtab:llama}
\end{subtable}
\end{table}

Results show that: firstly, MDIR is more robust than baselines such as REEF and HuRef. Even at 3x noise, MDIR retains more than 93\% of maximum possible trace value, while naive cosine similarity and CKA drops significantly to around 0.3-0.5. This demonstrates that MDIR preserves structural signal far better than standard similarity metrics under strong perturbation.

More importantly, noise destroys model functionality before detection fails. At 1x noise, both models suffer catastrophic performance collapse (MMLU drops to near-random). Yet, MDIR still reports a normalized trace greater than 0.99, indicating detection remains highly confident. This implies that if the adversary attempts to bypass detection via noise injection, they would also significantly degrade model performance.

\textbf{Retraining under a Different Tokenizer.} As we claim that our method remains efficient even with changed tokenizers, there lacks real-world examples of pretraining under a changed tokenizer. To prove the effectiveness of MDIR in such settings, we select Qwen2.5-0.5B as the base model (hidden dimension $n=896$), which originally uses tied embedding/unembedding, with Qwen2Tokenizer (vocabulary size 151,665, padded to 151,936). We discard its original embeddings, initialize a new embedding matrix of size $50304 \times 896$ with all zeros (trainable due to non-zero gradient), and continually pretrain the model on a DCLM subsample \citep{li2025datacomplmsearchgenerationtraining} (the first 100 files of shard 0 \citep{dclm_baseline_shard_2024}, 12.05B tokens), tokenized using the GPT-NeoX tokenizer (vocabulary size 50,277, padded to 50,304). All other weights are inherited from Qwen2.5-0.5B, so we anticipate that MDIR detects model homology. All weights (not just embeddings) undergo bfloat16 precision training. The starting learning rate is $10^{-4}$ with no warm-up, followed by quadratic decay to 0. We monitor three key metrics throughout training: the trace $\max_P\Tr\left(P \cdot \mathrm{Ortho}(E'^{\text{T}}E)^{\text{T}}\right)$,
the $p$-value, and the number of correctly matched channels (i.e., number of 1's in the permutation matrix $P$ that matches the correct position). The results are summarized in the Table~\ref{tab:summary}.
\begin{table}[h]
\centering
\small
\caption{MDIR successfully detects homology between Qwen2.5-0.5B retrained under a different tokenizer and the base model.}
\label{tab:summary}
\begin{tabular}{rrcr}
\toprule
Tokens (B) & Trace & $p$-value & \# Channels \\
\midrule
1.64 & 94.5 & NS & 110 \\
3.28 & 179.6 & $10^{-4750}$ & 828 \\
4.82 & 254.6 & $10^{-11813}$ & 896 \\
6.55 & 296.1 & $10^{-16799}$ & 896 \\
8.19 & 316.8 & $10^{-19533}$ & 896 \\
9.83 & 324.3 & $10^{-20584}$ & 896 \\
11.47 & 325.8 & $10^{-20798}$ & 896 \\
12.05 & 325.9 & $10^{-20806}$ & 896 \\
\bottomrule
\end{tabular}
\end{table}

We observe that, at early training stages (1.64B tokens), the trace is low and the $p$-value is not significant (NS), indicating minimal alignment, since the new embeddings are still undertrained. By 3.28B tokens, the $p$-value plunges to $10^{-4750}$ (very significant), and 828 of 896 channels are correctly matched. This suggests that the vocabulary embeddings have almost recovered the structure, despite the new tokenizer. Starting around 4.82B tokens, All channel relationships are matched, and the $p$-value continues to go down steadily. This implies that MDIR remains effective even when pretraining with a changed tokenizer.

\subsection{Ablation Experiment}
To demonstrate that MDIR exclusively detects relevance in weights, rather than training data, we conduct an ablation experiment by initializing two models with different random seeds and training each on two distinct datasets, resulting in a total of 4 models.
The datasets are DCLM subsample \citep{li2025datacomplmsearchgenerationtraining} (first 100 files of shard 0 \citep{dclm_baseline_shard_2024}, 12.05B tokens), and OpenWebMath-ProX \citep{zhou2025programmingexampleliftingpretraining} (4.61B tokens \citep{owm_prox_2024}).
Both datasets are tokenized using the GPT-NeoX tokenizer \citep{black-etal-2022-gpt}.  
Models are configured with the \texttt{Qwen3ForCausalLM} \citep{huggingface_qwen3_impl} architecture, with 12 layers and an intermediate size of 1024,  
resulting in a total of 291 million parameters.  
They are initialized using HuggingFace \texttt{transformers}' default initialization range of 0.02, 
with random seeds $2$ and $3$, respectively.  
All models are trained with a learning rate of linear warmup to $0.002$ followed by quadratic decay to $0$,  
and batch size $8 (\text{GPUs}) \times 48 (\text{sequences}) \times 1024 (\text{length})$.

Visualization of the trace values for the embedding matrices between models initialized with the same seed reveal clear diagonal patterns,  
indicating strong weight similarity due to shared initialization.  
In contrast, visualization of the traces for models that did not share a seed show no significant outliers and no significance of $p$-values,
even though block-wise patterns are present for inner transformation matrices of the attention module.
This suggests that models trained on the same dataset may develop similar attention features but no substantial weight correlation.  Please see Figure~\ref{fig:seed} in Appendix~\ref{sec:matviz} for these visualizations.

\section{Conclusion}

We introduced MDIR, a novel SOTA method for detecting weight homology in large language models, 
leveraging matrix analysis and Large Deviation Theory to provide statistically rigorous results.  
MDIR operates directly on model weights (bypassing full inference) and can execute on low-resource devices, improving accessibility of the verification process. It can operate on model pairs with different tokenizers and even different layer counts, allowing detection of the provenance of upcycled or pruned models. Outputs can include specific weight relationships found in numerical or visual form, indicating what kinds of transformations were applied to a source model.

\textbf{Why Extreme $p$-values. } We produce $p$-values as small as $10^{-10^4}$ or lower, far beyond typical statistical thresholds.
This is not a numerical artifact, but an accurate assessment of the statistical probability that the weights could have come from unrelated initializations. Such strong evidence is a direct consequence of Large Deviation Theory (LDT) as applied to high dimensional data.
LDT characterizes tail probabilities via rate functions: $p \asymp \exp(-n^2 I(x))$, 
where $n$ is the model dimension and $I$ is some rate function quantifying deviation from the null. 
For modern LLMs with $n \gg 1$, this could lead to astronomically small $p$-values.
This reflects the fundamental difference between LLM-scale statistics (billions or even trillions of parameters) and classical statistical scenarios (hundreds of samples).  
Such values underflow in floating-point arithmetic but remain well-defined and computable via $\log p$.

\textbf{Limitation and Future Work.}
While MDIR demonstrates effectiveness in model homology detection across various scales, it currently struggles to study model \textit{phylogeny}, which involves understanding the evolutionary relationships among models. For instance, let $A, B, C$ be three homologous models. Is model $C$ more closely related to model $A$ or model $B$? How does the last common ancestor of $A$ and $B$ look like? Currently, our method cannot answer complex phylogenetic questions yet. We hope to inspire future work, potentially integrating multidisciplinary insights to better understand the "family trees" of language models.

\section*{Impact Statement}

We present MDIR, a novel method for detecting model homology, fostering accountability and transparency in the field of machine learning. Our work has important implications for both intellectual property protection and the ethical development of AI systems. 

We strongly uphold the principles of intellectual property rights and oppose any form of model plagiarism or unauthorized replication of models. While our method can be used to provide evidence of potential model copying by organizations or individuals, it is not designed to make direct accusations against any entity. The interpretation and handling of such findings should be conducted carefully and with appropriate legal and ethical considerations.

\bibliography{main}
\bibliographystyle{icml2026}

\newpage
\appendix
\onecolumn

\section{Matrix Visualizations}
\label{sec:matviz}
\begin{figure}[htbp]
    \centering
    \subfigure[Embedding]{
        \includegraphics[width=0.31\textwidth]{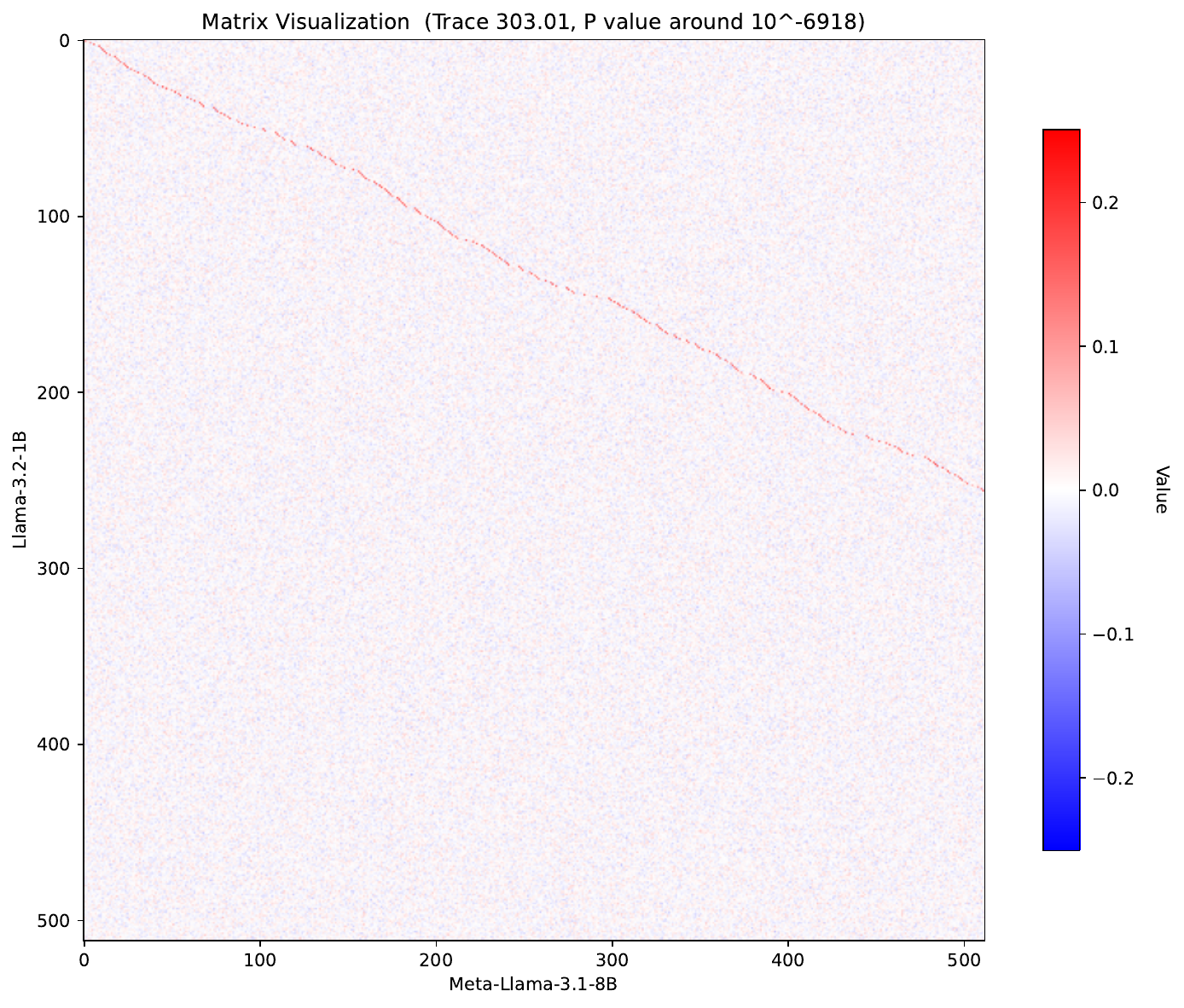}
    }
    \subfigure[Layer 3 Attention V]{
        \includegraphics[width=0.31\textwidth]{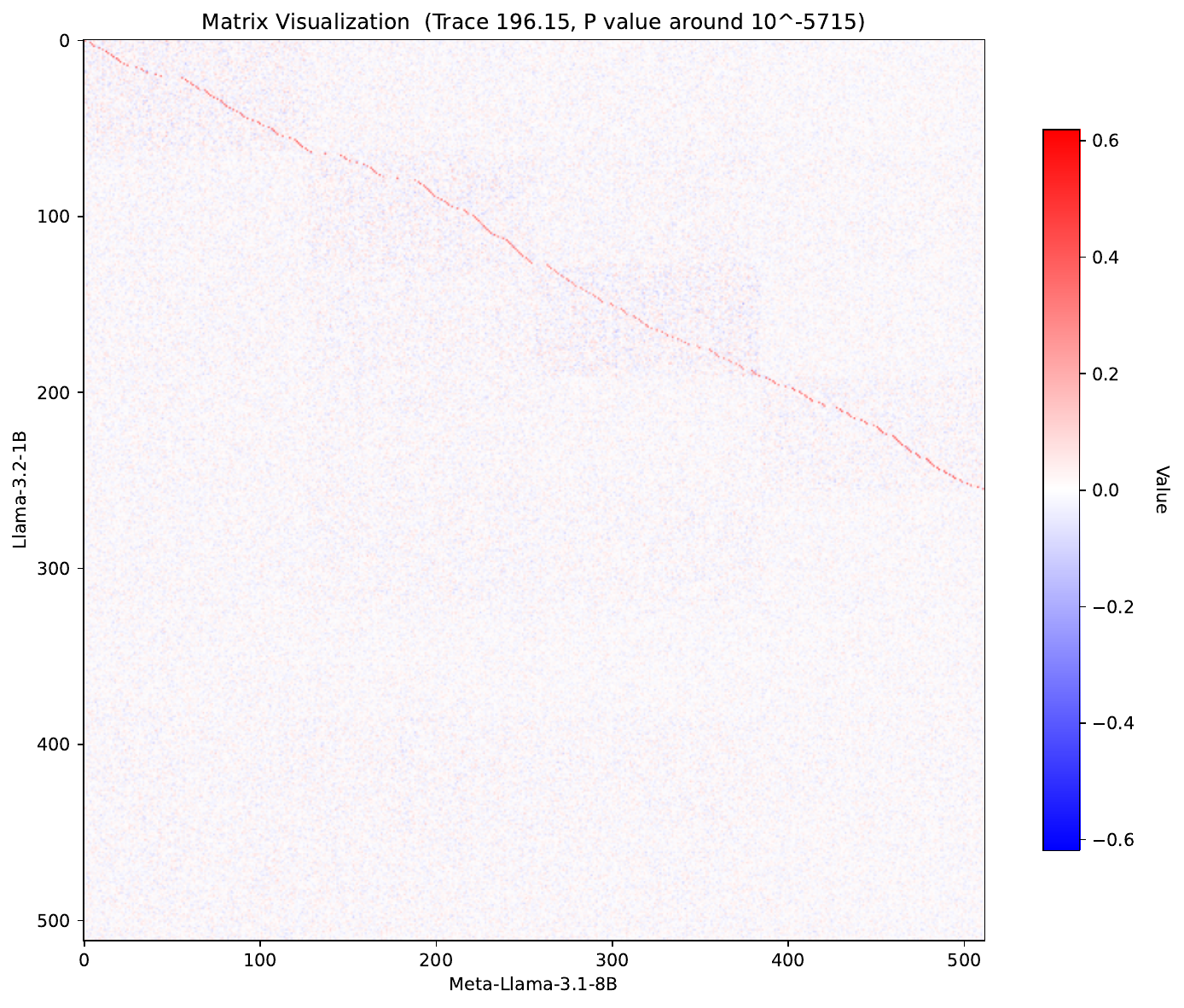}
    }
    \subfigure[Layer 3 MLP Up]{
        \includegraphics[width=0.31\textwidth]{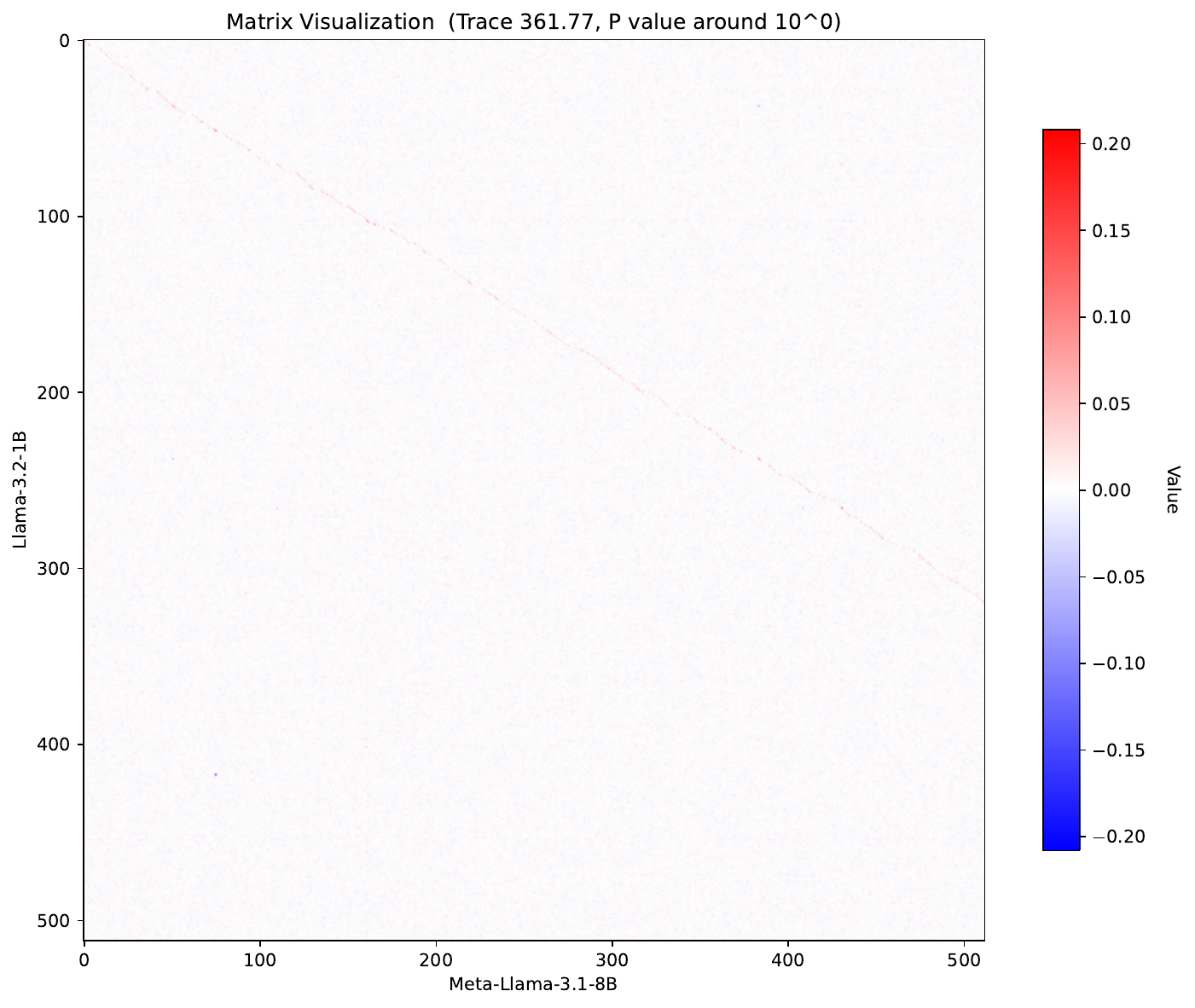}
    }
    \caption{MDIR suggests homology between Llama-3.1-8B and Llama-3.2-1B. yielding a $p$-value of $10^{-6,918}$. 
    For model pruning, the irregular oblique curves (the slope is approximately $1/2$, indicating that half of the channels are retained) can be clearly identified in $\tilde{U}$ from vocabulary as well as inner transformations in the attention module.}
    \label{fig:llama31_llama32_1b}
\end{figure}

\begin{figure}[htbp]
    \centering
    \subfigure[Embedding]{
        \includegraphics[width=0.31\textwidth]{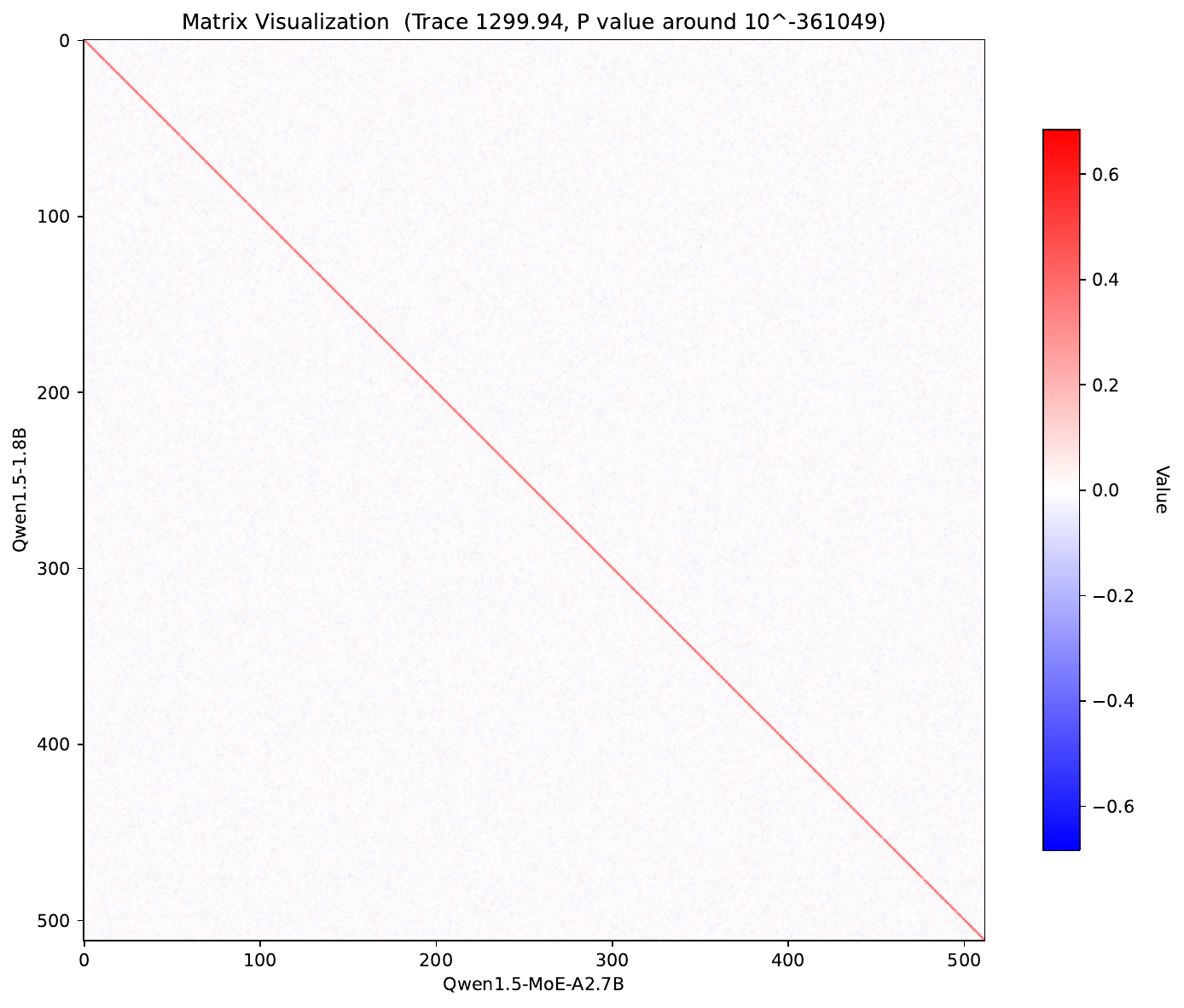}
    }
    \subfigure[Layer 9 Attention V]{
        \includegraphics[width=0.31\textwidth]{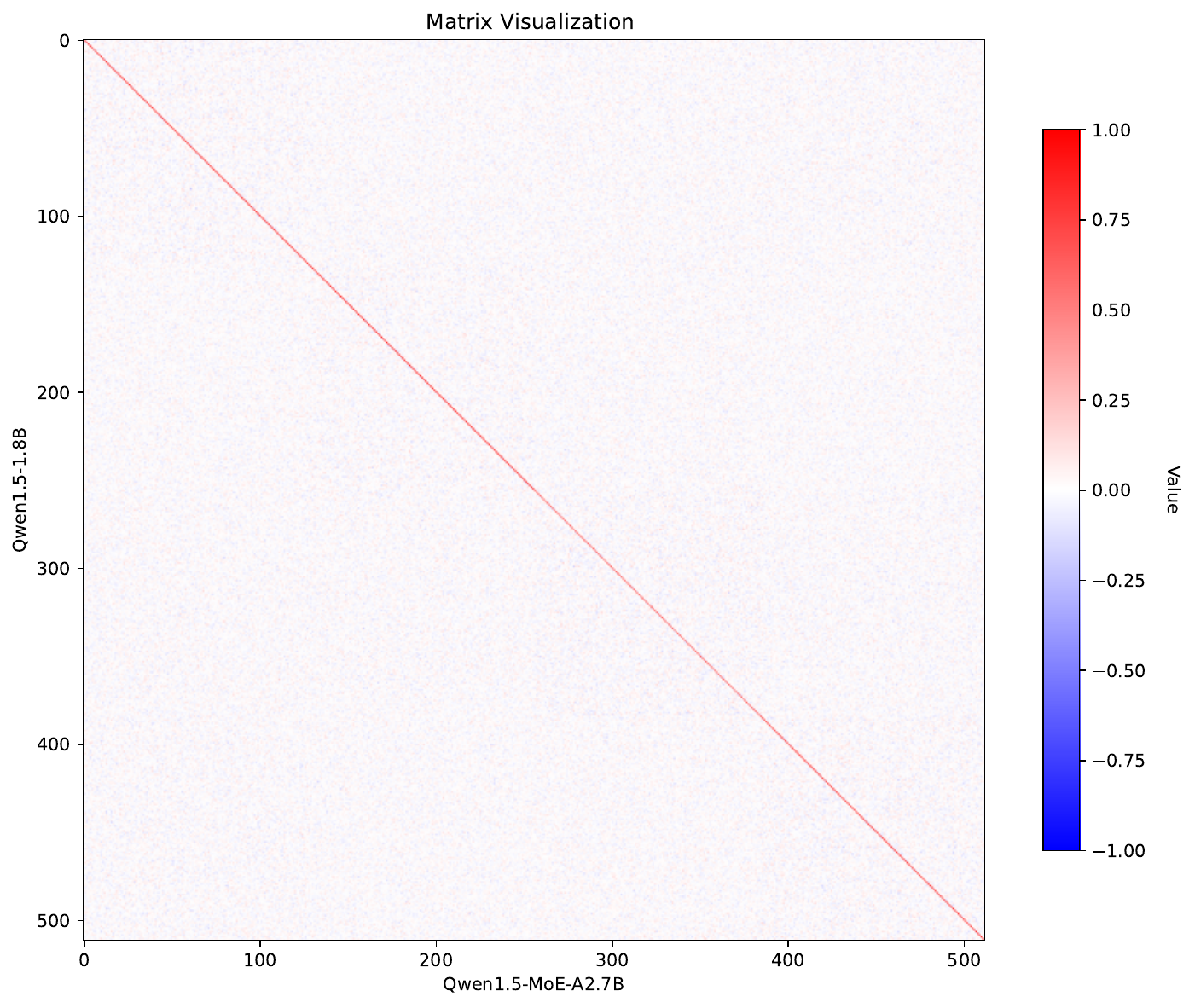}
    }
    \subfigure[Layer 9 MLP Up]{
        \includegraphics[width=0.31\textwidth]{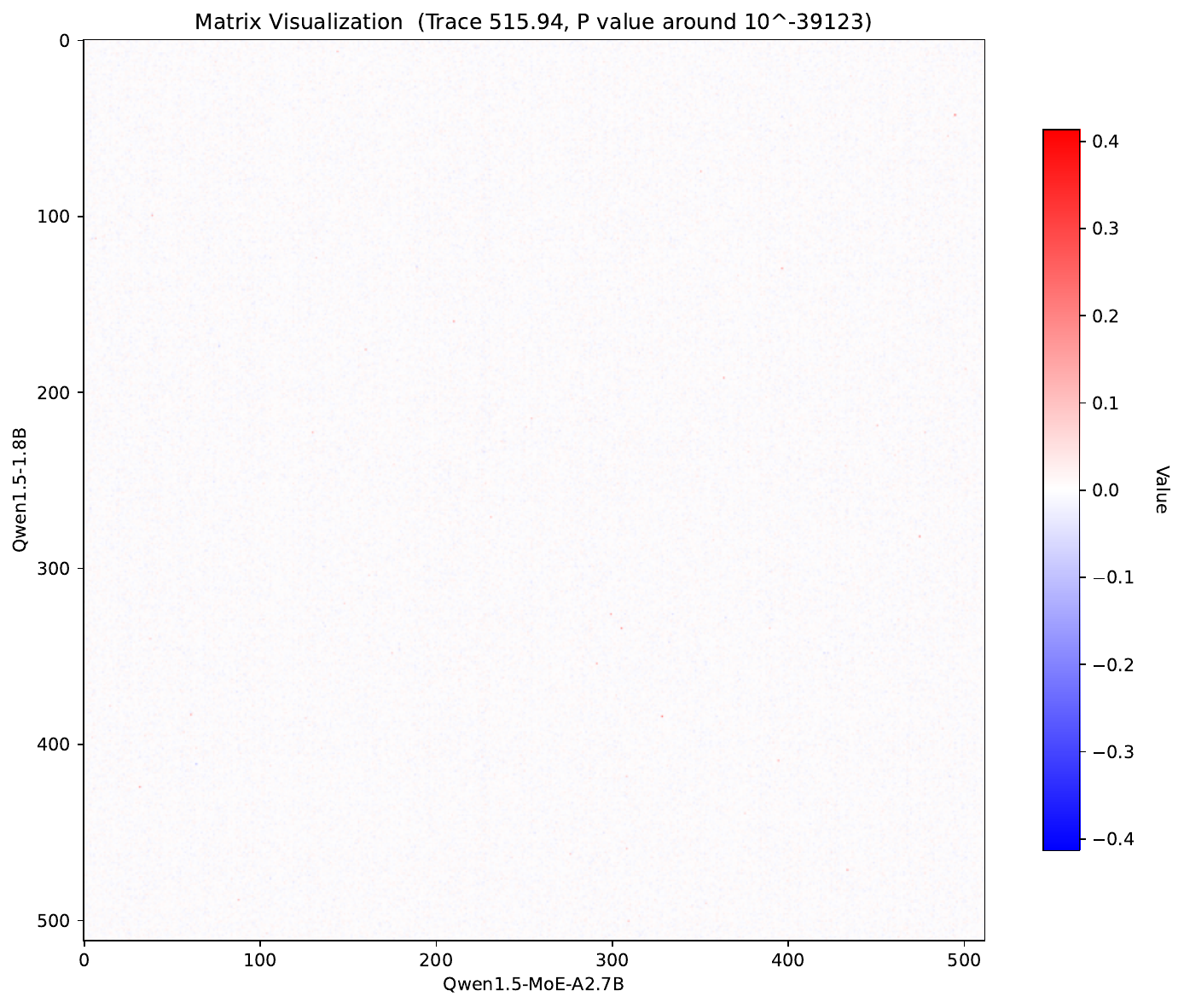}
    }
    \caption{For model upcycling, MDIR suggests homology between Qwen1.5-1.8B and Qwen1.5-MoE-A2.7B, yielding a $p$-value of $10^{-361,049}$. The diagonal patterns for vocabulary embedding and attention modules indicate that these modules are directly inherited from its predecessor, and show no evidence of permutation or channel reselection before the upscaling process.}
    \label{fig:qwen15_18}
\end{figure}

\begin{figure}[htbp]
    \centering
    \subfigure[Embedding]{
        \includegraphics[width=0.31\textwidth]{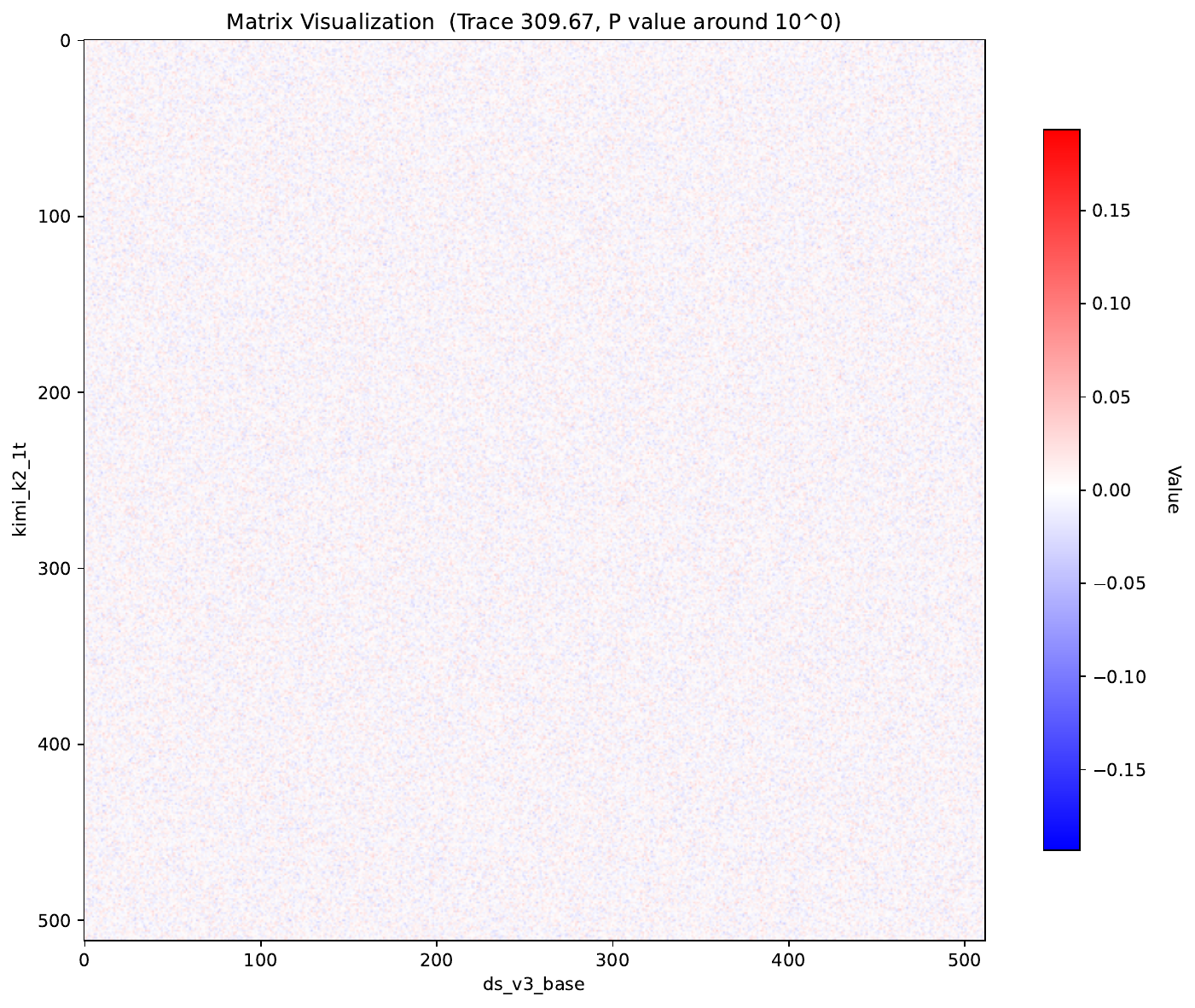}
    }
    \subfigure[Layer 0 Attention O]{
        \includegraphics[width=0.31\textwidth]{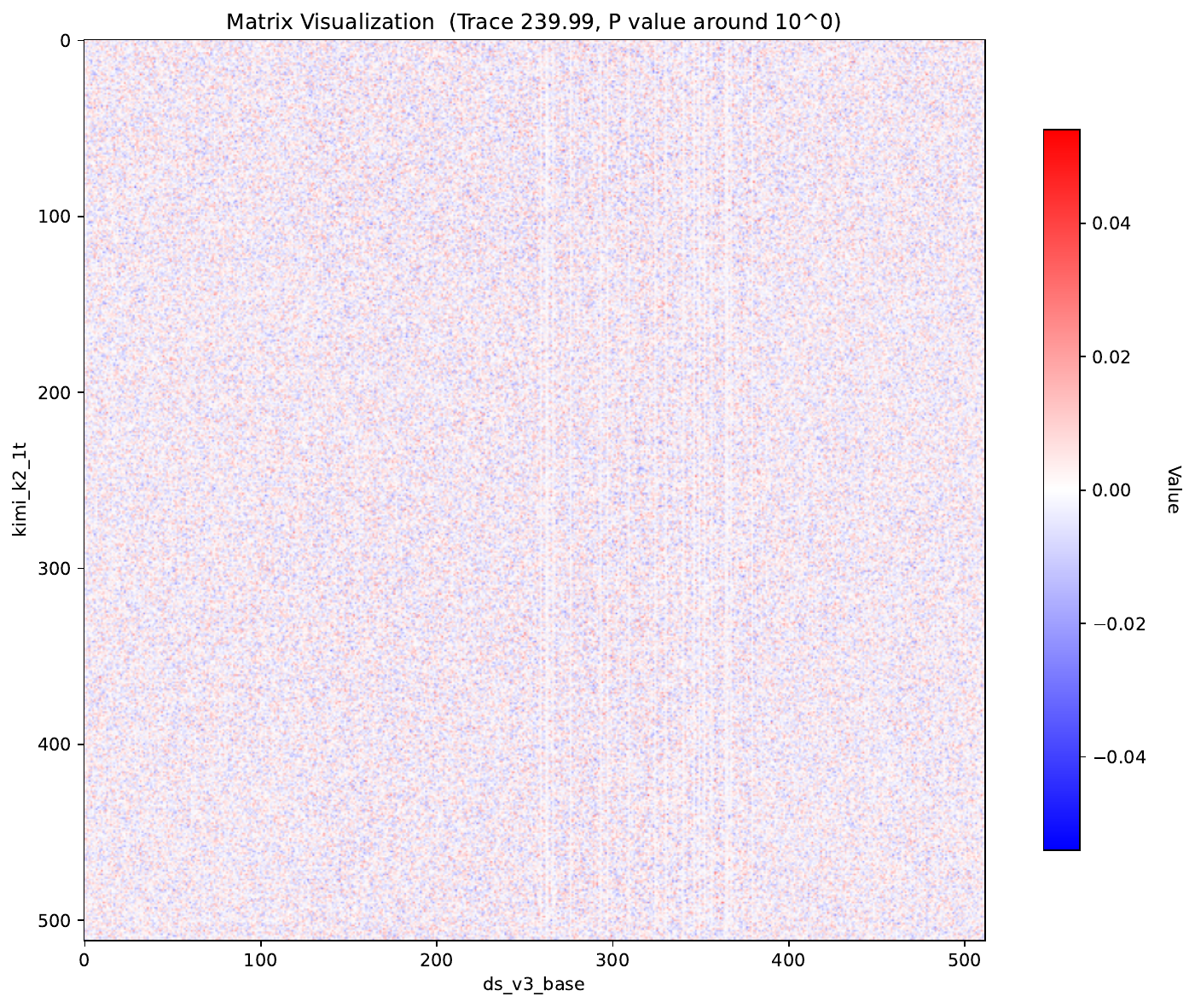}
    }
    \subfigure[Layer 0 MLP Up]{
        \includegraphics[width=0.31\textwidth]{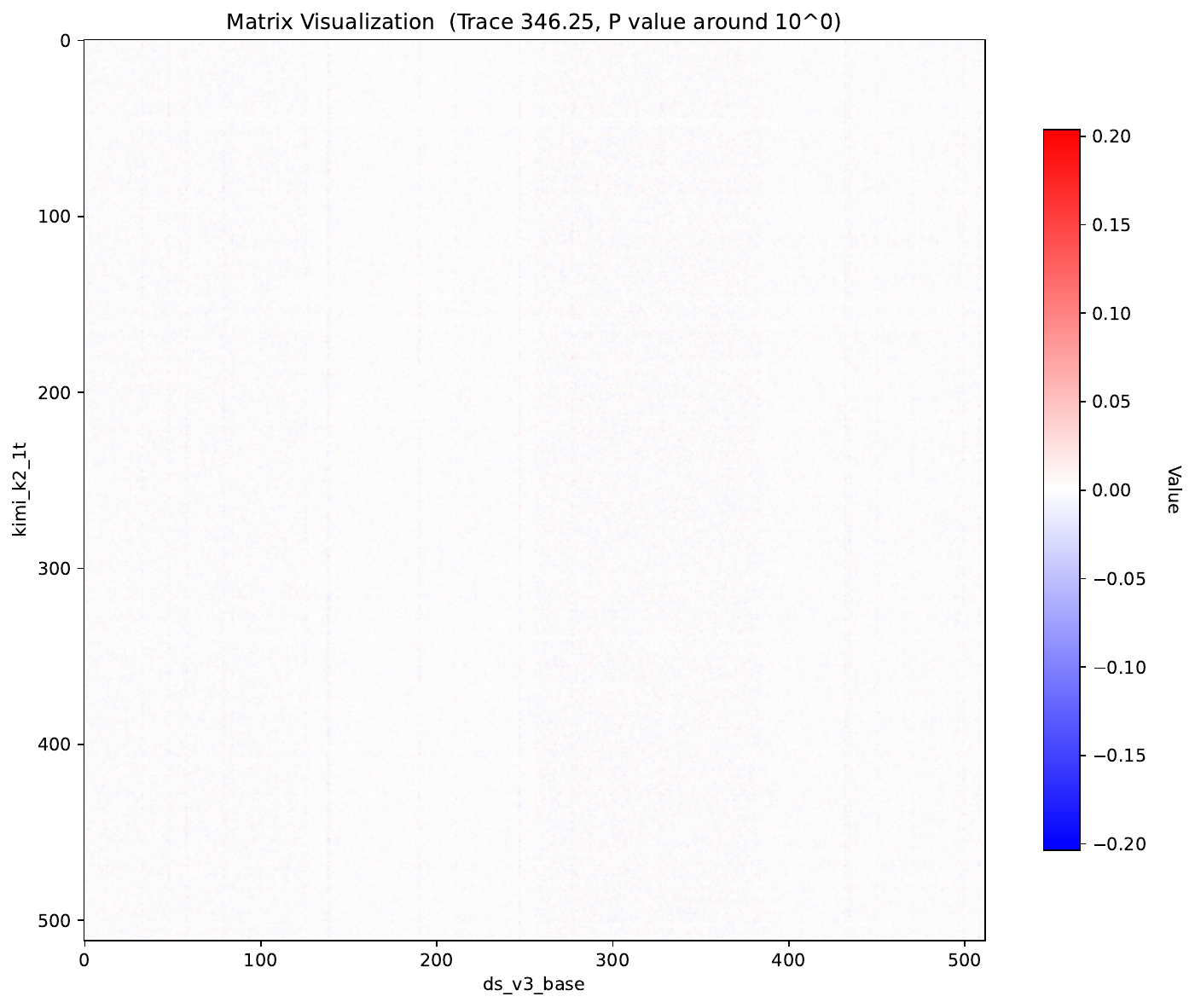}
    }
    \caption{For independently developed models, MDIR detects no statistically significant homology between DeepSeek-V3-Base and Kimi-K2-Instruct, with no clear pattern or statistically significant $p$-value observed. }
    \label{fig:ds_kimi}
\end{figure}

We visualize some matrices for several representative cases in Figures \ref{fig:llama31_llama32_1b}, \ref{fig:qwen15_18} and \ref{fig:ds_kimi}.

\begin{figure}[htbp]
    \centering
    \subfigure[Seed 2, embed]{
        \includegraphics[width=0.23\textwidth]{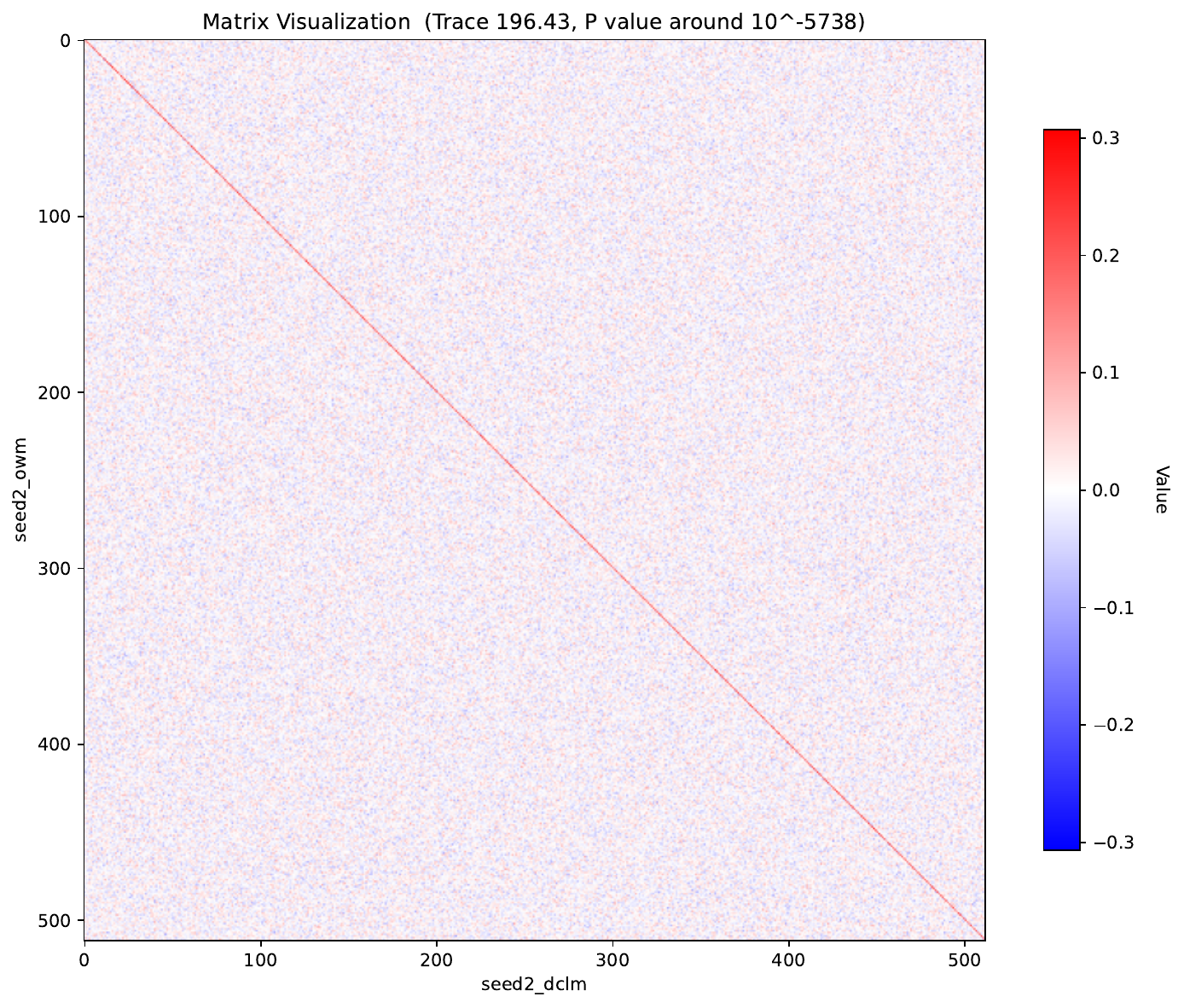}
    }
    \subfigure[Seed 3, Attn V]{
        \includegraphics[width=0.23\textwidth]{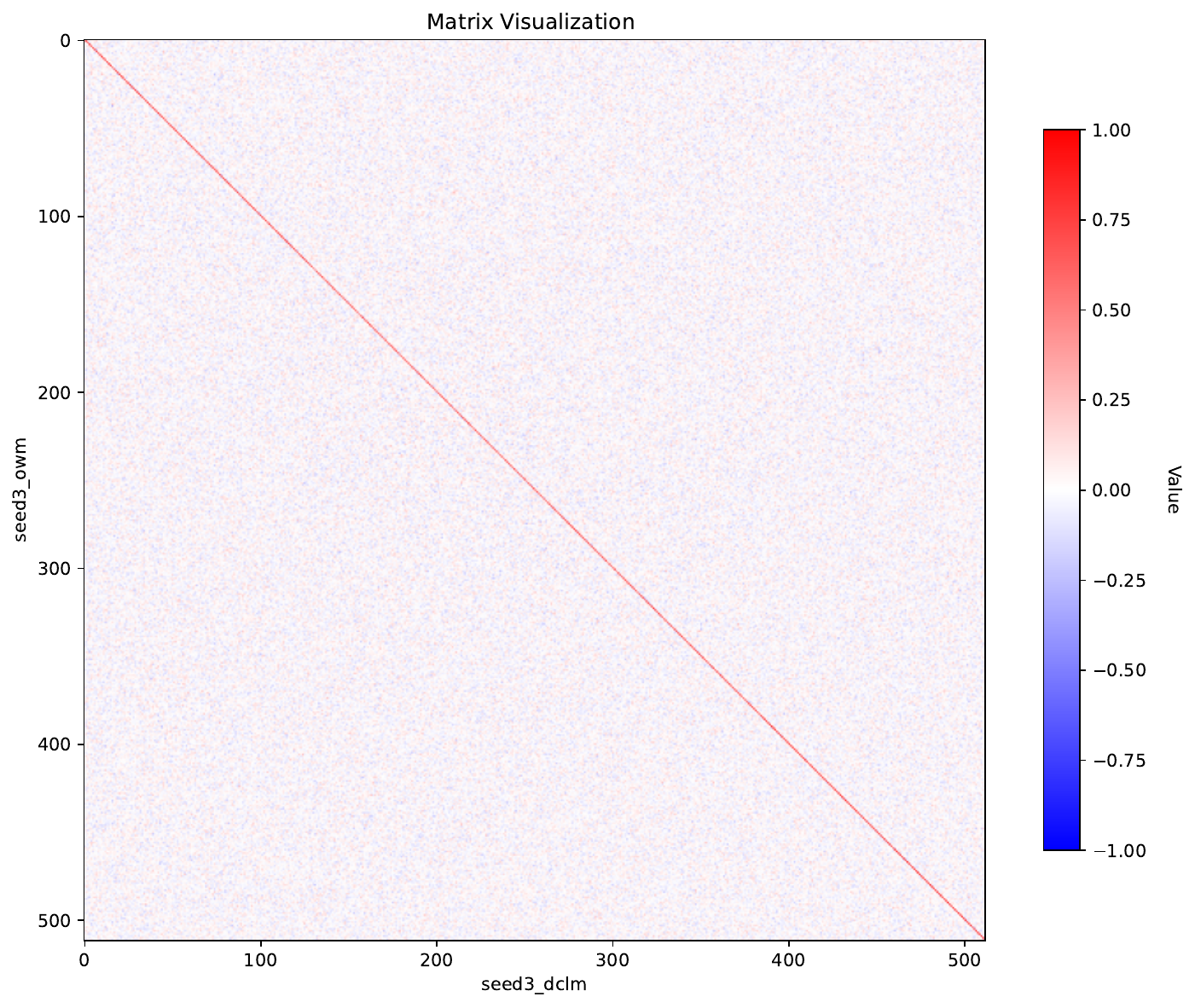}
    }
    \subfigure[OWM, embed]{
        \includegraphics[width=0.23\textwidth]{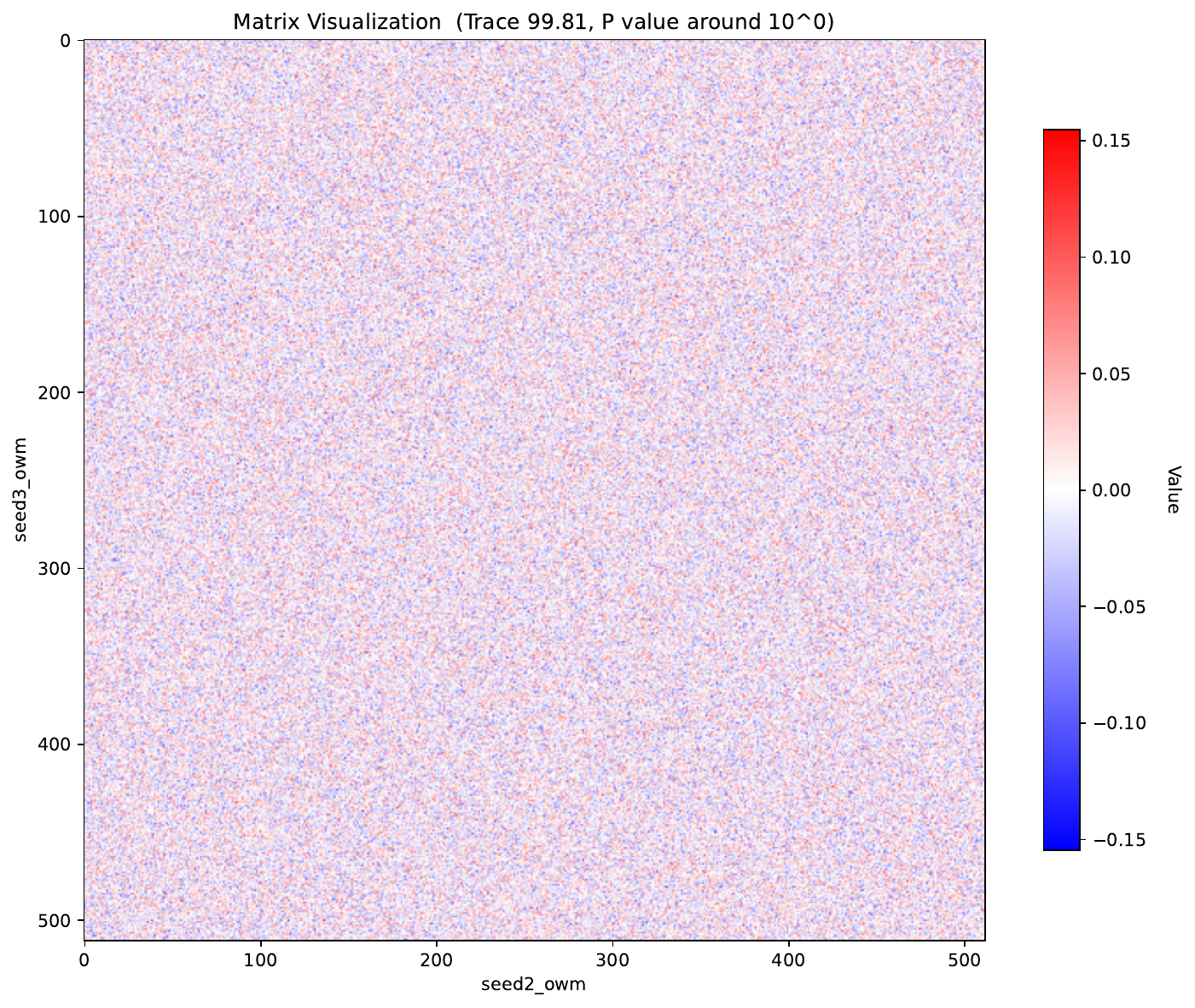}
    }
    \subfigure[DCLM, Attn V]{
        \includegraphics[width=0.23\textwidth]{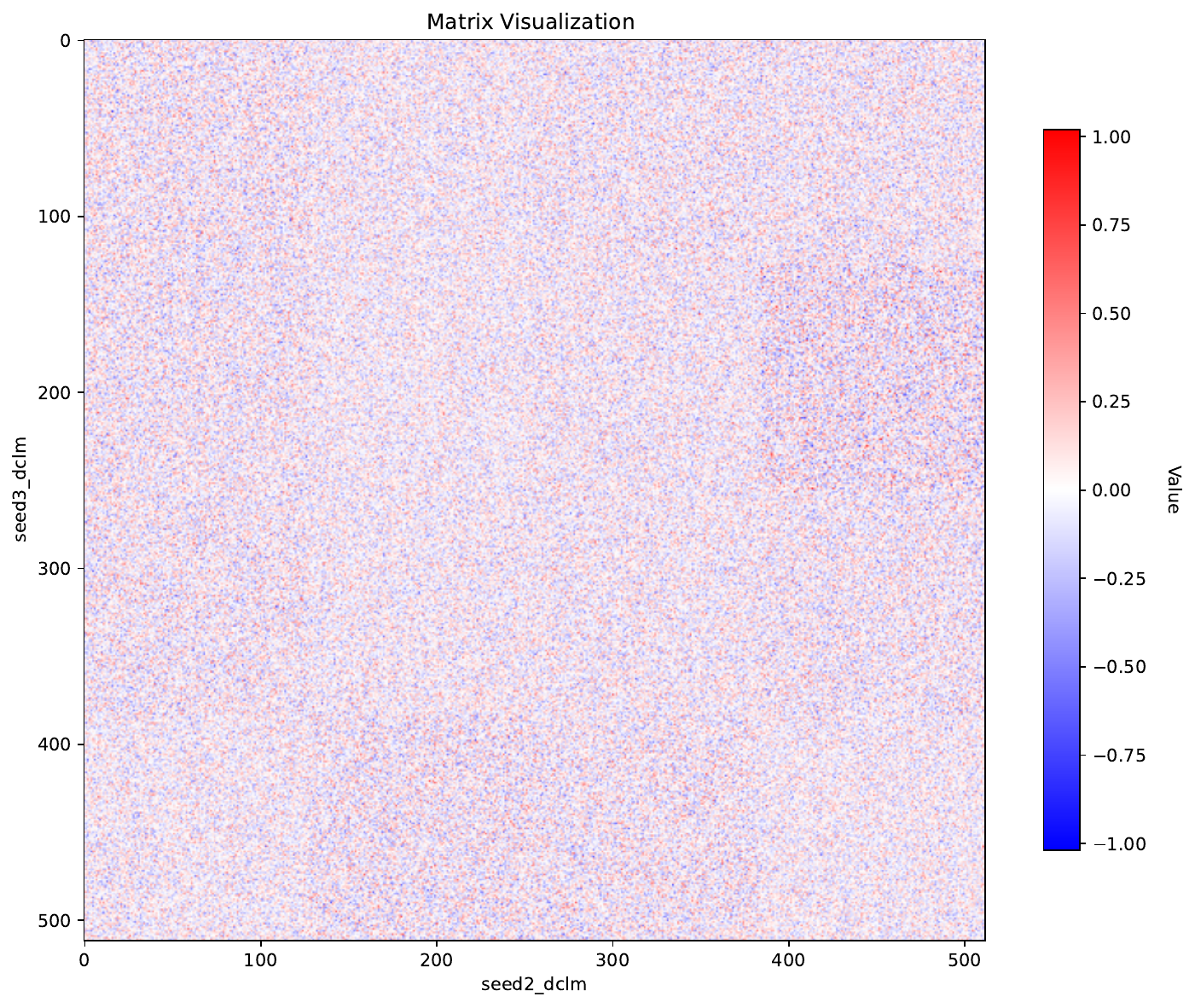}
    }
    \caption{MDIR only identifies relationship between models initialized with the same seed.}
    \label{fig:seed}
\end{figure}

\section{Mathematical Background}
Our work primarily builds upon the foundation of matrix analysis. To avoid excessive technicality or verbosity, we only introduce the key tools and properties used in this paper, potentially omitting or abbreviating proofs.  
Interested readers are encouraged to refer to \citet{horn2012matrix} for a comprehensive overview of matrix analysis.

\subsection{Matrix Analysis: Singular Value and Polar Decompositions}

\textbf{Singular Value Decomposition.} The singular value decomposition (SVD) of a matrix \( A \in \R^{m \times n} \) is given by
\[
A = U S V^{\text{T}},
\]
where \( U \in \R^{m \times m} \) and \( V \in \R^{n \times n} \) are orthogonal matrices, 
and \( S \in \mathbb{R}^{m \times n} \) is a diagonal matrix with non-negative singular values on the diagonal:
\[
\sigma_1 \geq \sigma_2 \geq \cdots \geq \sigma_r > 0 \quad (r = \mathrm{rank}(A)).
\]
We denote \(\sigma_i(A)\) as the \(i\)-th singular value of \(A\).

\textbf{Polar Decomposition.} The polar decomposition of \(A\) has two common but distinct forms: 
the left decomposition \( A = PW \) and the right decomposition \( A = WQ \),
where \( P = (AA^{\text{T}})^{1/2} \) and \( Q = (A^{\text{T}}A)^{1/2} \) are symmetric positive semidefinite matrices,
and \( W \in \R^{m \times n} \) (with orthonormal columns) is shared between both decompositions.
We define the \textit{orthogonal part} of \(A\) as \(\mathrm{Ortho}(A) := W\). 
When \(A\) is full-rank, the orthogonal part of \(A\) is unique.

\textbf{Connection to SVD.} The orthogonal part \(W\) in the polar decomposition can be obtained from the SVD:
\[
W = U V^{\text{T}},
\]
where \(U\) and \(V\) are derived from the SVD of \(A\).
The symmetric factors satisfy \( P = U S U^{\text{T}} \) and \( Q = V S V^{\text{T}} \).
However, when \(A\) is not invertible (with many singular values either exactly zero or close to zero), the computation of \(W = U V^{\text{T}}\) becomes ill-conditioned.

\textbf{SVD and Spectral Calculus.} Assume \(A \in \R^{m \times n}\) has SVD \(A = U S V^{\text{T}}\). Then:
\[
AA^{\text{T}} = U (SS^{\text{T}}) U^{\text{T}}, \quad A^{\text{T}}A = V (S^{\text{T}} S) V^{\text{T}}.
\]
For any polynomial function \(f \in \R[x]\) and \(g(x) = x f(x^2)\), we have:
\[
\begin{aligned}
    & f(AA^{\text{T}})A \\
    &= U f(SS^{\text{T}}) U^{\text{T}} U S V^{\text{T}} = U f(SS^{\text{T}}) S V^{\text{T}} \\
    &= U g(S) V^{\text{T}} \\
    &= U S f(S^{\text{T}} S) V^{\text{T}} = U S V^{\text{T}} V f(S^{\text{T}} S) V^{\text{T}} \\
    &= A f(A^{\text{T}} A).
\end{aligned}
\]
In fact, \(g(S)\) can represent any odd polynomial. For any odd function \(G\), we may find a polynomial \(f\) such that \(f(x)\) agrees with \(G(x)/x\) on all nonzero diagonal entries of \(SS^{\text{T}}\).  
In particular, setting \(G(x) = \mathrm{sign}(x)\), this spectral calculus yields the most ``sensible'' orthogonal part of \(A\).  
We adopt the methods from \citet{amsel2025polarexpressoptimalmatrix} for a fast and relatively accurate implementation.

\textbf{Trace Maximization Property.} The orthogonal part \(W\) solves the optimization problem:
\[
\max_{\{ X | X^{\text{T}} X = I \}} \left( \Tr(A X^{\text{T}}) \right).
\]
The maximum value is the sum of all the singular values of \(A\), i.e., \(\Tr(P) = \sum_{i=1}^r \sigma_i\).

\textbf{Orthogonal Invariance of Singular Values.} The singular values \(\{\sigma_i\}\) of \(A\) are invariant under both left and right orthogonal transformations:
\[
\sigma_i(UAV) = \sigma_i(A), \quad \forall U, V \text{ orthogonal and } i = 1, \cdots, \min(m,n).
\]
Thus, any function depending on these singular values remains invariant under orthogonal transformations. Examples include:

\begin{itemize}
    \item \textbf{Frobenius Norm}: \( \lVert A \rVert_F = \sqrt{\sum_i \sigma_i^2(A)} \);
    \item \textbf{Spectral Norm}: \( \lVert A \rVert_S = \sigma_1(A) \);
    \item \textbf{Ky Fan $k$-Norm}: \( \lVert A \rVert_{KF} = \sum_{i=1}^k \sigma_i(A) \);
    \item \textbf{Schatten $p$-Norm} \citep{schatten_cross_norms}: \( \lVert A \rVert_p = \sqrt[p]{\sum_i \sigma_i^p(A)} \).
\end{itemize}

Any combination of these functions also remains invariant under orthogonal transformations. These functions may serve as preliminary indicators for detecting model similarity.

\textbf{Orthogonal Matrices and RMSNorms Commute.} For any orthogonal matrix \(U\), we have:
\[
\mathrm{RMSNorm}(x)U = \mathrm{RMSNorm}(x U),
\]
for any nonzero vector \(x \in \R^n\) (row vector). Moreover, all transformations satisfying this property \((\mathrm{RMSNorm}(\cdot))U = \mathrm{RMSNorm}((\cdot)U)\) are orthogonal transformations.

To prove this, note that \(\mathrm{RMSNorm}(y)\) is always a constant multiple (\(\sqrt{n}\)) of a unit vector when \(y \neq 0\).  
Thus, \(U\) maps all unit vectors to unit vectors.  
By linearity, this implies \(\lVert xU \rVert = \lVert x \rVert\) and \(x U U^{\text{T}} x^{\text{T}} = x x^{\text{T}}\).  
Taking \(x\) over all eigenspaces of \(U U^{\text{T}}\), we see that \(1\) is the only eigenvalue of \(U U^{\text{T}}\).  
Hence, \(U U^{\text{T}} = \bm{1}_n\), and \(U\) is orthogonal.

\subsection{Large Deviation Theory}
Our research extensively involves random orthogonal matrices, 
particularly focusing on traces of such matrices. 
To obtain statistically meaningful $p$-values, 
traditional statistical $p$-tests become useless for our case,
due to the interdependence of the elements within an orthogonal matrix. 
Instead, we rely on large deviation theory for deriving the $p$-value.
For a comprehensive exposition of this theory, please refer to \citet{DemboZeitouni1998} and \citet{anderson2010introduction}.

\section{Preservation under Scaling and Permutation}
\label{sec:preservation_under_scaling_and_permutation}
Following Equation~\ref{eqmainformula}, we let $E$ and $E'$ denote the vocabulary embeddings of model $A$ and $B$ respectively. 
We demonstrate that the final results remain invariant when an adversary applies both scaling and permutation to the matrix $E'$. 
Specifically, we consider the transformed embedding $E'' = \alpha E' P'$, where $P' \in \mathrm{Perm}(\mathrm{n})$ is a permutation matrix, and $\alpha > 0$ is a scaling factor.
\begin{thm}
Define the trace and the corresponding optimal permutation respectively as follows:
\[
\begin{aligned}
    T(E, E') &= \max_{P \in \mathrm{Perm}(\mathrm{n})} \Tr\left(P \cdot \mathrm{Ortho}\left(E'^{\text{T}} E\right)^{\text{T}}\right), \\
    \Pi(E, E') &= \argmax_{P \in \mathrm{Perm}(\mathrm{n})} \Tr\left(P \cdot \mathrm{Ortho}\left(E'^{\text{T}} E\right)^{\text{T}}\right).
\end{aligned}
\]
Suppose an adversary applies a simultaneous scaling and permutation transformation to $E'$, yielding $E'' = \alpha E' P'$, where $P' \in \mathrm{Perm}(\mathrm{n})$ is an arbitrary permutation matrix, and $\alpha > 0$ is a positive scaling coefficient. In this case, the following properties hold:
\[
T(E, E'') = T(E, E'), \quad \Pi(E, E'') = P'^{\text{T}} \Pi(E, E').
\]
Thus, such adversarial modifications are futile.
\end{thm}
\begin{proof}
We first analyze the trace $T(E, E'')$:
\[
\begin{aligned}
    T(E, E'') &= \max_{P \in \mathrm{Perm}(\mathrm{n})} \Tr\left(P \cdot \mathrm{Ortho}\left((E''^{\text{T}} E)^{\text{T}}\right)\right) \\
    &= \max_{P \in \mathrm{Perm}(\mathrm{n})} \Tr\left(P \cdot \mathrm{Ortho}\left(((\alpha E' P')^{\text{T}} E)^{\text{T}}\right)\right) \\
    &= \max_{P \in \mathrm{Perm}(\mathrm{n})} \Tr\left(P \cdot \mathrm{Ortho}\left(\alpha E^{\text{T}} E' P'\right)\right).
\end{aligned}
\]
Since scaling by $\alpha > 0$ does not affect the orthogonal factor in polar decomposition ($\mathrm{Ortho}$), we can simplify further:
\[
\begin{aligned}
    T(E, E'') &= \max_{P \in \mathrm{Perm}(\mathrm{n})} \Tr\left(P \cdot \mathrm{Ortho}\left((E^{\text{T}} E') P'\right)\right) \\
    &= \max_{P \in \mathrm{Perm}(\mathrm{n})} \Tr\left(P \cdot \mathrm{Ortho}(E^{\text{T}} E') P'\right) \\
    &= \max_{P \in \mathrm{Perm}(\mathrm{n})} \Tr\left(P' P \cdot \mathrm{Ortho}\left(E^{\text{T}} E'\right)\right).
\end{aligned}
\]
Let $R = P' P$. Since $P'$ is a fixed permutation matrix, as $P$ ranges over all permutations in $\mathrm{Perm}(\mathrm{n})$, so does $R$. Substituting $R$ into the expression, we obtain:
\[
\begin{aligned}
    T(E, E'') &= \max_{R \in \mathrm{Perm}(\mathrm{n})} \Tr\left(R \cdot \mathrm{Ortho}\left(E^{\text{T}} E'\right)\right) \\
    &= \max_{R \in \mathrm{Perm}(\mathrm{n})} \Tr\left(R \cdot \mathrm{Ortho}\left((E'^{\text{T}} E)^{\text{T}}\right)\right) \\
    &= T(E, E').
\end{aligned}
\]
Next, we examine the optimal permutation $\Pi(E, E'')$. Recall that the trace achieves its maximum when $R = \Pi(E, E')$. From the substitution $R = P' P$, it follows that:
\[
P' P = \Pi(E, E') \! \implies \! P = P'^{-1} \Pi(E, E') = P'^{\text{T}} \Pi(E, E').
\]
Thus, the optimal permutation for $E''$ is given by:
\[
\Pi(E, E'') = P'^{\text{T}} \Pi(E, E').
\]
This completes the proof.
\end{proof}

\section{Uniform Distribution across Haar Measure}
\label{app:uniform}
\begin{thm}
Let $E'$ be an $m \times n$ random matrix initialized with i.i.d. Gaussian entries $\sim \mathcal{N}(0, \sigma^2)$, 
and let $E$ be an $m \times n$ matrix such that $\mathrm{rank}(E) = n$ ($m \ge n$ because vocabulary size is always large). 
We know that $E'^{\text{T}}E$ is almost surely full-rank. Define the function $\mathrm{Ortho}(X) = X(X^{\text{T}} X)^{-1/2}$, 
which gives the orthogonal factor in the polar decomposition of $X$. Then $\mathrm{Ortho}(E'^{\text{T}}E)$ is uniformly distributed across $\mathrm{O}(n)$ with respect to the Haar measure.
\end{thm}

\begin{proof}
We proceed as follows:

\textbf{Lemma. } For any orthogonal matrix $U \in O(n)$ and any $n \times n$ matrix $X$, 
\[
\mathrm{Ortho}(UX) = U \, \mathrm{Ortho}(X).
\]

\textbf{Proof. } By definition, $\mathrm{Ortho}(X) = X(X^T X)^{-1/2}$. Substituting $UX$ into the formula:
\[
\mathrm{Ortho}(UX) = (UX)\left((UX)^\text{T}(UX)\right)^{-1/2} = UX\left(X^{\text{T}}U^{\text{T}}UX\right)^{-1/2} =  UX\left(X^{\text{T}}X\right)^{-1/2} = U \, \mathrm{Ortho}(X).
\]

\textbf{Lemma. }  If $U \in O(n)$ is any orthogonal matrix, then $UE'^{\text{T}}$ has the same distribution as $E'^{\text{T}}$.

\textbf{Proof. } The rows of $E'$ are i.i.d. Gaussian random vectors $\sim \mathcal{N}(0, \sigma^2I)$, which are isotropic. For each column, we have $E'_j U^{\text{T}} \overset{d}{=} E'_{j}$ and i.i.d. for different $j=1, \cdots, m$. Therefore, $E'U^{\text{T}} \overset{d}{=} E'$ and $UE'^{\text{T}} \overset{d}{=} E'^{\text{T}}$.

We now combine the lemmas to prove the main result: Let $U \in O(n)$ be any orthogonal matrix. Consider the distribution of $\mathrm{Ortho}(E'^{\text{T}}E)$ under the transformation induced by $U$. Using first lemma:
\[
\mathrm{Ortho}(U E'^{\text{T}}E) = U \, \mathrm{Ortho}(E'^{\text{T}}E).
\]
From the second lemma, since $U E'^{\text{T}}$ has the same distribution as $E'^{\text{T}}$, it follows that $U E'^{\text{T}}E$ has the same distribution as $E'^{\text{T}}E$. Consequently, $\mathrm{Ortho}(U E'^{\text{T}}E)$ has the same distribution as $\mathrm{Ortho}(E'^{\text{T}}E)$. 

This invariance under left multiplication by any orthogonal matrix $U$ implies that $\mathrm{Ortho}(E'^{\text{T}}E)$ is uniformly distributed on $O(n)$ with respect to the Haar measure.
\end{proof}

\section{Inner Transformations in the Attention Module}
\label{sec:inner_transformations_attention}

Our main algorithm covers a basic set of inner transformations within attention. However, a more complete implementation might take into account the following complex relationship that could be contained within $W_Q, W_K, W_V, W_O$. We present the extended version here, while the algorithm listing that we use in practice (see Appendix~\ref{sec:algorithm}) excludes some features like doing separate head permutation analysis.

The inner transformations for layer $\ell \ (1 \le \ell \le L)$ is summarized as
\[
\begin{aligned}
    &W_{Q,\ell} = \mu_\ell \cdot P_{1,\ell} \otimes P_{2,\ell} \otimes S_\ell, \quad \\
    &W_{K,\ell} = \mu^{-1}_\ell \cdot P_{1,\ell} \otimes S_\ell, \\
    &W_{V,\ell} = \sum_{v=1}^{\mathrm{NumKVHeads}} (\mathbf{1}_{v, \sigma_\ell(v)} \otimes H_{\ell,v}), \\
    &W_{O,\ell} = \sum_{v=1}^{\mathrm{NumKVHeads}} (\mathbf{1}_{v, \sigma_\ell(v)} \otimes P_2 \otimes H_{\ell,v}), \\ 
    &(1 \le \ell \le L)
\end{aligned}
\]

where $\otimes$ denotes the Kronecker product of matrices, $\mu \neq 0$ is a scalar, $P_1 \in \mathrm{Perm}(\mathrm{NumKVHeads})$ is a permutation matrix over $\mathrm{NumKVHeads}$ channels (and $\sigma$ is the corresponding permutation such that $\sum \mathbf{1}_{v, \sigma(v)} = P_1$),  
$P_2 \in \mathrm{Perm}(\mathrm{QueriesPerHead})$ is a permutation matrix over $\mathrm{QueriesPerHead}$ queries,  
$S \in \mathrm{diag}(\pm 1, \cdots, \pm 1) \in \mathrm{M}_{\mathrm{HeadDim}}(\mathbb{R})$, and $H \in \mathrm{O}(\mathrm{HeadDim}, \mathbb{R})$ is an arbitrary orthogonal matrix.

To ensure compatibility with both QK-norm \citep{henry2020querykeynormalizationtransformers} and RoPE \citep{su2023roformerenhancedtransformerrotary}, $S$ is restricted to diagonal matrices with entries $\pm 1$:  
$S \in \mathrm{diag}(\pm 1, \cdots, \pm 1).$

The totally invariant group $G$ can be generated by these elements:
\[ 
\begin{aligned}
&U, \ \ (P_{1,\ell})_{1 \le \ell \le L},\ \  (P_{2,\ell})_{1 \le \ell \le L},\ \  (S_{\ell})_{1 \le \ell \le L}, \\
&(H_{\ell,v})_{1 \le \ell \le L, 1\le v \le \mathrm{NumKVHeads}}.
\end{aligned}
\]
Where $U \in \mathrm{O}(\mathrm{EmbDim})$ is the component that contributes the most dimension ($(\mathrm{EmbDim}^2 - \mathrm{EmbDim})/2$) to this group.

Each component satisfies the constraints outlined in the paper:
\begin{itemize}
    \item $U$ represents the outer transformation, which is shared across all layers and belongs to the orthogonal group $\mathrm{O}(\mathrm{EmbDim})$.
    \item $(P_{1,\ell})_{1 \le \ell \le L}$ and $(P_{2,\ell})_{1 \le \ell \le L}$ are permutation matrices corresponding to attention heads and queries respectively.
    \item $(S_{\ell})_{1 \le \ell \le L}$ are diagonal matrices with entries $\pm 1$ to ensure compatibility with Rotary Positional Embeddings.
    \item $(H_{\ell,v})_{1 \le \ell \le L, 1\le v \le \mathrm{NumKVHeads}}$ are general invertible matrices applied to individual attention heads.
\end{itemize}

In other words, $G$ is the Cartesian product of these components:
\[
G = ( U )  \times \prod_{\ell=1}^L \left( (P_{1,\ell}) \times (P_{2,\ell}) \times (S_\ell) \times \prod_{v=1}^{\mathrm{NumKVHeads}} (H_{\ell,v}) \right)
\]

\begin{figure}[htbp]
  \centering
  \includegraphics[width=1.0\linewidth]{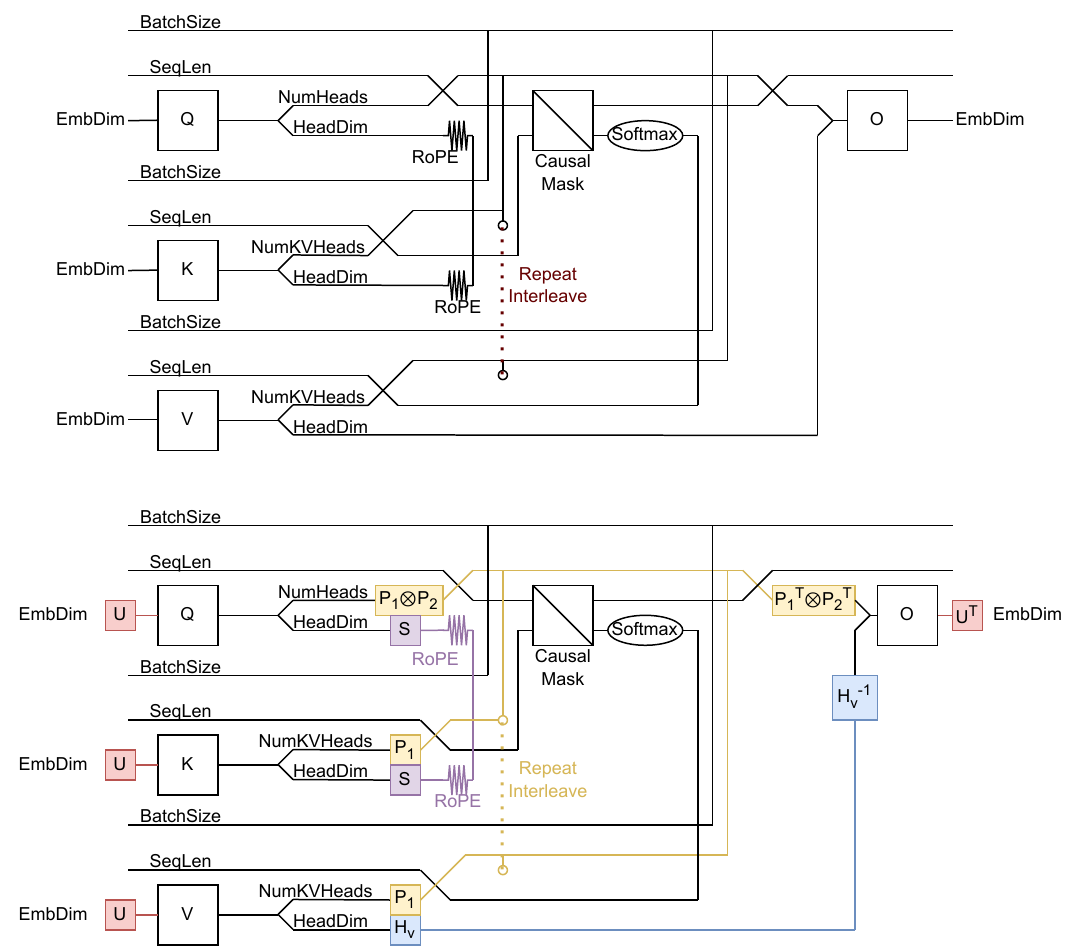}
  \caption{A Penrose notation of a model with Grouped Query Attention architecture (model $A$, up) and an illustration of an adversary $B$ under application of invariant transformations (down). Bias terms are omitted, and RMSNorms are not explicitly shown.}
  \label{fig:gqa}
\end{figure}

We summarize the GQA architecture and invariant transformations in Figure \ref{fig:gqa}, illustrated using Penrose tensor notation, which compactly encodes the tensor contractions and symmetry operations underlying the GQA module, and may serve a visual illustration why the inner transformations take the form above.

\section{Solving All Transformations}
If the significance of the $p$-value has already been determined from the $p$-values for vocabulary embeddings,  
this step is entirely optional. However, if the no significant $p$-value is observed at this stage,  
we cannot yet rule out the possibility of model homology. This may occur when an adversary used a general orthogonal matrix for obfuscation.  
Additionally, interested readers may wish to determine the exact relationship between two models,  
potentially with different architectures.

Since $P = \arg\max_P \Tr(P \tilde{U}^{\text{T}})$, we proceed as follows:  
If $P$ is reliably identified as a permutation matrix ($p < p_0$, where $p_0$ is some threshold), set $U := P$; otherwise, set $U := \tilde{U}$.

\textbf{Solving the Relationship between Layers.}
In practical scenarios, the number of layers for two models are not necessarily the same. 
It is important to first determine the relationship between the layers of both models via solving a maximal ordered matching problem.
We select a representative matrix (for example, attention K or V) for each layer $i$ of model $A$ and layer $j$ of model $B$.
For each pair $(i,j)$ we do the process below and obtain a similarity measure matrix of shape $L \times L'$. 
Solving the maximal ordered matching for this matrix will give the relationship between all layers. See Section \ref{sec_comparison}
for a specific example how this works.

\textbf{Solving the Transformations in the Attention Module.}  
We now solve for the inner transformations $W_Q$, $W_K$, $W_V$, and $W_O$ based on the heuristic $U$.  

Our objectives are:
\[
\min_{W_Q \text{ orthogonal}} \lVert U Q W_Q - Q' \rVert_F^2, \ \  \min_{W_K \text{ orthogonal}} \lVert U K W_K - K' \rVert_F^2,
\]
for $Q$ and $K$. By the trace maximization property, the solutions for $W_Q$ and $W_K$ are given by:
\[
W_Q = \lambda \mathrm{Ortho}(Q^{\text{T}} U^{\text{T}} Q'), \quad W_K = \mu \mathrm{Ortho}(K^{\text{T}} U^{\text{T}} K'),
\]
where the layer-wise similarity measure can be calculated as
\[ t_\square = \texttt{linear\_sum\_assignment}(W_\square), \quad \square \in Q, K.\]
Also, scaling coefficients $\lambda$ and $\mu$ are computed as:
\[
\lambda = \frac{\lVert Q' \rVert_F}{\lVert Q \rVert_F}, \quad \mu = \frac{\lVert K' \rVert_F}{\lVert K \rVert_F}.
\]

For $W_V$ and $W_O$,  
we may apply the same method to compute $\mathrm{Ortho}(V^{\text{T}} U^{\text{T}} V')$ and $\mathrm{Ortho}(O U^{\text{T}} O'^{\text{T}})$.  
However, since $W_V$ and $W_O$ may involve general invertible transformations, our method does not guarantee recovering the exact transformation.
solving the general case without additional assumptions is challenging.  
We leave this problem for future work.

\textbf{Solving the MLP.}  
At this point, there is only one matrix left to solve: the permutation of intermediate neurons, denoted by $P$.  
The solution is given as follows:
\[
\begin{aligned}
U_X &= \mathrm{Ortho}(X^{\text{T}} U^{\text{T}} X'), \quad X \in \{\mathrm{Gate}, \mathrm{Up}, \mathrm{Down}\}; \\
P &= \argmax_{P \in \mathrm{Perm}(\mathrm{IntermediateDim})} \Tr \left( P \left( U_{\mathrm{Gate}} + U_{\mathrm{Up}} + U_{\mathrm{Down}} \right)^{\text{T}} \right).
\end{aligned}
\]

Typically, we expect the three solutions $\arg\max \Tr \left( P U_X^{\text{T}} \right)$, for $X \in \{\mathrm{Gate}, \mathrm{Up}, \mathrm{Down}\}$,  
to yield the same permutation. However, computing the orthonormal part for intermediate matrices (which often have 14,000–20,000 rows) is computationally expensive.  
Adding the three terms together would triple the computation cost. If the noise level is tolerable,  
we may simply select one of them:
\[
P = \argmax_{P \in \mathrm{Perm}(\mathrm{IntermediateDim})} \Tr \left( P U_{\mathrm{Up}}^{\text{T}} \right).
\]

\section{Overall Algorithm}
\label{sec:algorithm}

We summarize our algorithm in three steps: embedding analysis (Algorithm~\ref{alg:embed}), layer correspondence matching (Algorithm~\ref{alg:match}), and layer analysis (Algorithm~\ref{alg:layer}). Embedding analysis is the first step for MDIR homology detection. In most cases not involving a sophisticated adversary, this is all that is needed to make a strong detection. For model pairs with different numbers of layers, one can then run MDIR layer correspondence matching. This results in a $p$-value for every pair of layers $(i, j)$ which can be used to determine which layers match. Finally, MDIR layer analysis can be run on each pair of matching layers.

\begin{algorithm}[htbp]
\caption{MDIR layer correspondence matching}
\label{alg:match}
    \begin{algorithmic}[1]
    \For{Layer $i \in [L_A]$}
        \For{Layer $j \in [L_B]$}
            \For{$(X, X') \in \{V\}$ of $(A_i,B_j)$}
                \State $U_X \gets \mathrm{Ortho}(X^{\text{T}} U_E^{\text{T}} X')$ 
                \State $P_X \gets \arg\max_{P \in \mathrm{Perm}(\mathrm{D_{attn}})} \Tr(P U_X^{\text{T}})$
                \State $p_X[i, j] \gets L \cdot \mathrm{PValue}(P_X, U_X, \mathrm{D_{attn}})$
            \EndFor
        \EndFor
    \EndFor
    \end{algorithmic}
\end{algorithm}

\begin{algorithm}[htbp]
\caption{MDIR layer analysis}
\label{alg:layer}
    \begin{algorithmic}[1]
    \For{$(X, X') \in \{Q, K, V, Up\}$ of $(A,B)$}
        \State $U_X \gets \mathrm{Ortho}(X^{\text{T}} U_E^{\text{T}} X')$
        \State $P_X \gets \arg\max_{P \in \mathrm{Perm}(\mathrm{D_X})} \Tr(P U_X^{\text{T}})$
        \State $p_X \gets L \cdot \mathrm{PValue}(P_X, U_X, \mathrm{D_X})$
    \EndFor
    \end{algorithmic}
\end{algorithm}

\section{Estimation of $p$-value via Large Deviation Theory}
\label{appendix_estimate_p}
Let $\tilde{U}$ be a random orthogonal matrix, distributed uniformly according to the normalized Haar measure. 
We aim to estimate the following function, especially its long-tail behavior, for $c > 0$:
\[
f(c) = \mathbb{P}\left[\Tr(\tilde{U}) \geq c\right].
\]

The distribution of $f(c)$ is a well-studied problem in random matrix theory.  
\citet{DiaconisShahshahani1994} proved that $\Tr(\tilde{U}) \to \mathcal{N}(0, 1)$ in distribution.  
Later, \citet{Johansson1997} showed that the convergence of $\Tr(\tilde{U})$ to $\mathcal{N}(0, 1)$ is exponential under the total variation distance:
\[
\mathrm{TV}(f - (1 - \Phi)) \leq \exp(-\alpha n), \quad \text{for some } \alpha > 0,
\]
where $\Phi(x)$ is the cumulative distribution function of the standard normal distribution.

It is known that $(1 - \Phi)(x)$ has the following asymptotic behavior for large $x$:
\[
(1 - \Phi)(x) \approx \frac{1}{\sqrt{2\pi} x} \exp\left(-\frac{x^2}{2}\right), \quad (x \gg 1).
\]
However, when both $n$ and $x$ are large, $\exp(-\alpha n)$ remains significantly larger than $\frac{1}{\sqrt{2\pi} x} \exp\left(-\frac{x^2}{2}\right)$.  
Thus, it is inappropriate to use $(1 - \Phi)$ as the asymptotics of $f(c)$.

To study the tail behavior of $f(c)$, we leverage tools from Large Deviation Theory.  
Since $n$ is typically large and the embedding dimension $n$ is usually an even number in most models,  
we assume $n = 2m$ without loss of generality.  
Additionally, we assume $\det(\tilde{U}) = 1$, or equivalently $\tilde{U} \in \mathrm{SO}(2m)$,  
since this introduces only an infinitesimal difference in the thermodynamic limit ($n \to \infty$).  
These assumptions simplify our analysis while preserving accuracy.

\begin{thm}
    Let $A \in \mathrm{SO}(2m)$ be uniformly distributed according to the normalized Haar probability measure, and let $0 < r \leq 1/2$.  
    The probability 
    \[
    P(r, m) = \mathbb{P} \left[\frac{1}{2m} \Tr(A) \geq r \right]
    \]
    satisfies the following large deviation principle:
    \[
    \lim_{m \to \infty} \frac{-\log P(r, m)}{2 m^2 r^2} = 1,
    \]
    or equivalently,
    \[
    P(r, m) \asymp \exp(-m^2 I(r)),
    \]
    where 
    \[
    I(r) = 2 r^2
    \]
    is the good rate function.

    For $r > 1/2$, a faster decay rate can be achieved: $I(r) > 2 r^2$.
    \label{thm:ldp}
\end{thm}

Before we delve into the proof, it is worth discussing the challenges we encountered. 
We are dealing with the Circular Real Ensemble (CRE, named after \citet{PhysRevLett.104.147001}), 
which has a formulation distinct from traditional circular ensembles, such as
Circular Unitary Ensembles (CUE) or Circular Orthogonal Ensembles (COE) \citep{Dyson1962}. 
These ensembles play important roles in both random matrix theory and condensed matter physics.

One of the problems most similar to ours is the Gross-Witten Ensemble \citep{GrossWitten1980,gamboa2017sumruleslargedeviations}, 
which concerns the large deviations from the typical value of $\mathrm{Re} \Tr (U)$ as $N \to \infty$
(where $U \in \mathrm{U}(N)$ is a random unitary matrix). Our problem can be regarded as a real orthogonal variant of Gross-Witten Ensemble, 
but this is undocumented in previous literature.

Unitary matrices possess many elegant properties. 
One of them is the Cayley transformation, defined as $\phi(z) := \mathrm{i}(1+z)/(1-z)$, 
which maps the punctured unit circle $\mathbb{T} \setminus \{1\}$ to the real line $\mathbb{R}$.
When the Cayley transformation is applied to a unitary matrix, it sends unit eigenvalues to the real line,
thereby producing a Hermitian matrix (i.e., $\phi(U)$ is Hermitian for $U \in \mathrm{U}(N)$).
However, under the Cayley transformation, an orthogonal matrix is transformed into 
a purely imaginary, anti-symmetric matrix, which is of little interest. 
In the field of deep learning, complex values rarely appear, and we shall focus on the Circular Real Ensemble 
in our proof.

The circular ensemble is closely related to thermodynamics and large deviation theory. 
We refer to Mehta's book \textit{Random Matrices} \citep{Mehta2004}, 
and a comprehensive introduction can be found in Chapter 12 of it.

\subsection{Proof of Theorem \ref{thm:ldp}}
\begin{proof}
    If $A$ is uniformly distributed according to the Haar measure in $\mathrm{SO}(2m)$ (a manifold of real dimension $m(2m-1)$), 
    all eigenvalues of $A$ lie on the unit circle, 
    and complex eigenvalues form paired conjugates, possibly accompanied by several $+1$ and $-1$. 
    
    When $\det A = 1$, the product of complex eigenvalues yields $+1$, and there are almost surely no $-1$ or $+1$ eigenvalues. 

    Denote by $\{e^{i\theta_k}, e^{-i\theta_k} : k=1,2,\cdots, m\}$ the eigenvalues of $A$. It is a classical result (see \citet{weyl2016classical,Girko1985}) 
    that the phases $(\theta_k)_k$ obey the distribution characterized by the following probability density:
    \[p(\theta_1, \cdots, \theta_m)\mathrm{d}\theta_1 \cdots\mathrm{d}\theta_m  = C\prod_{1\leq k<j\leq m}(\cos \theta_k - \cos \theta_j)^2\mathrm{d}\theta_1 \cdots\mathrm{d}\theta_m,\]
    and the trace $\Tr(A)$ is the sum of all eigenvalues:
    \[ \Tr(A) = 2\sum_{i=1}^m \cos(\theta_i).\]
    By substitution of variables, let $t_i = \cos(\theta_i) \in [-1,1]$, and $\mathrm{d}t_i / \sqrt{1-t_i^2}= \mathrm{d}\theta_i$. We study the substituted distribution:
    \[p(t_1, \cdots, t_m)\mathrm{d}t_1 \cdots\mathrm{d}t_m  = C' \prod_{1\leq k<j\leq m}(t_k - t_j)^2 \cdot \prod_{1\leq i \leq m} (1-t_i^2)^{-1/2} \mathrm{d}t_1 \cdots\mathrm{d}t_m.\]
    Taking the logarithm, we have
    \[-\log p(t_1, \cdots, t_m) = \sum_{1\leq k<j\leq m} \left(-2 \log|t_k - t_j| \right) + \sum_{1\leq i \leq m} \left(\frac{\log (1-t_i^2)}{2}\right) + C_0.\]
    This has a clear thermodynamical interpretation: Consider $m$ interacting particles located on the interval $[-1, 1]$, with coordinates $t_1, \cdots, t_m$. 
    The energy $E(t_1, \cdots, t_m)$ is the sum of the following two kinds of potentials:
    \begin{enumerate}
        \item Repelling force: for particles $k$ and $j$, their interaction potential is $-2 \log|t_k - t_j|$, meaning that they repel each other according to the 2-dimensional Coulomb law;
        \item External field: particles are attracted to the boundary points $-1$ and $+1$, with the potential $\sum_{1\leq i \leq m} \left(\frac{\log (1-t_i^2)}{2}\right)$. 
    \end{enumerate}
    Inspired by Chapter 12 of \citet{Mehta2004} and also \citet{TOUCHETTE20091}, 
    we define the Canonical Ensemble of these $m$ particles as follows, which admits a partition function:
    \[
    \begin{aligned}
        Z(t_1, \cdots, t_m) &= \int_{-1}^{1} \cdots \int_{-1}^{1} \exp({-\beta E(t_1, \cdots, t_m)}) \mathrm{d}t_1 \cdots\mathrm{d}t_m, \\
    \end{aligned}
    \]
    where
    \[
    \begin{aligned}
    E(t_1, \cdots, t_m) 
    &= \sum_{1\leq k<j\leq m} \left(-2 \log|t_k - t_j| \right) + \sum_{1\leq i \leq m} \left(\frac{\log (1-t_i^2)}{2}\right) \\
    &= -\log p(t_1, \cdots, t_m) - C_0.\\ 
    \end{aligned}
    \]
    We set the thermodynamic beta to be $\beta = 1/(k_B T) = 1$ because we do not study temperature changes. 

    We are interested in the probability:
    \[
    P(r, m) = \mathbb{P}\left[\frac{1}{2m} \Tr(A) \geq r\right].
    \]
    Using the representation $\Tr(A) = 2 \sum_{k=1}^m \cos(\theta_k) = 2 \sum_{k=1}^m t_k$, this becomes:
    \[
    P(r, m) = \mathbb{P}\left[\frac{1}{m} \sum_{k=1}^m t_k \geq r\right].
    \]
    Let \[\mu_m = \frac{1}{m} \sum_{k=1}^m \delta_{t_k}\] be the empirical measure of the eigenvalue distribution. 
    In the thermodynamic limit $m \to \infty$, $\mu_m$ should converge weakly to an equilibrium measure $\mu$ that minimizes the free energy functional $F(\mu)$. 

    We now solve the exact form of $F(\mu)$. We refer to Theorem 2.1 of the paper \citep{Eichelsbacher_2011},
    where we are dealing with the case $\theta = 1$, $\kappa = 1$, and $w_m (x) = (1-x^2)^{1/(2m)}$, where $w_m (x) \to 1$
    in the thermodynamic limit $m \to \infty$. This suggests that the interaction term dominates, and the external field term
    is negligible when $m$ is large. 

    The form of $F(\mu)$ is therefore given by
    \[
    F(\mu) = \iint_{[-1, 1]^2} \left( \log \frac{1}{|x-y|} \right) \mathrm{d} \mu(x) \mathrm{d} \mu(y),
    \]
    with the rate function
    \[
    \tilde{I}(\mu) = \iint_{[-1, 1]^2} \left( \log \frac{1}{|x-y|} \right) \mathrm{d} \mu(x) \mathrm{d} \mu(y) - c,
    \]
    where $c = \inf_\mu F(\mu)$. Note that the rate function $\tilde{I}(\mu)$ is defined for the probability measure $\mu$, not yet 
    ready for our $I(r)$. 
    
    From Corollary 2.2 of the paper \citep{Eichelsbacher_2011}, it suffices to solve the following two variational problems:
    \[ \inf_\mu F(\mu); \quad \inf_{\mu:\  \int x\mathrm{d} \mu \ge r} F(\mu). \]

    To solve these two problems, we parametrize $\mu$ using Chebyshev polynomials, with the additional assumption that 
    $\mu$ is absolutely continuous with respect to the Lebesgue measure $\mathrm{d}x$ 
    and normalized Chebyshev measure $\mathrm{d}x/(\pi \sqrt{1-x^2})$. 

    We suppose that $\mu$ is parametrized by the following series:
    \[ \mu(x) = \frac{1}{\pi \sqrt{1-x^2}} \sum_{i=0}^{\infty} a_i T_i(x),\]
    where $T_i$ is the $i$-th Chebyshev polynomial of the first kind, and $F(\mu)$ now becomes a quadratic form of these coefficients $\{a_i\}$.
    An additional constraint must not be overlooked: $\mu$ is a probability measure, and
    \[ \int_{-1}^{1} \mathrm{d} \mu = \frac{T_0(x)}{\pi \sqrt{1-x^2}} \sum_{i=0}^{\infty} a_i T_i(x) \mathrm{d} x = a_0 = 1.\] 

    It is known that $\log |x-y|$ has the Chebyshev expansion \citep{SloanStephan1992,MasonHandscomb2002}:
    \[ \log |x-y| = -\log 2 - \sum_{n=1}^{\infty} \frac{2}{n} T_n(x) T_n(y),\]
    so
    \[
    \begin{aligned}
        F(\mu) &= \iint_{[-1,1]^2} \left( \log 2 + 
        \sum_{n=1}^{\infty}\frac 2 n T_n(x) T_n(y) \right) \frac{1}{\pi^2 \sqrt{1-x^2}\sqrt{1-y^2}} \\
        &\quad \quad \quad \left(\sum_{i=0}^{\infty} a_i T_i(x)\right) \left(\sum_{i=0}^{\infty} a_i T_i(y)\right) \mathrm{d} x \mathrm{d}y \\
        &= \log 2 \cdot \iint_{[-1,1]^2} \frac{1}{\pi^2 \sqrt{1-x^2}\sqrt{1-y^2}} \mathrm{d} x \mathrm{d}y \\
        &\quad + \iint_{[-1,1]^2} \left(
        \sum_{n=1}^{\infty}\frac 2 n T_n(x) T_n(y) \right) \frac{1}{\pi^2 \sqrt{1-x^2}\sqrt{1-y^2}} \\
        &\quad \quad \quad \quad \left(\sum_{i=1}^{\infty} a_i T_i(x)\right) \left(\sum_{i=1}^{\infty} a_i T_i(y)\right) \mathrm{d}x \mathrm{d}y \\
        &= \log 2 \cdot \iint_{[-1,1]^2} \frac{1}{\pi^2 \sqrt{1-x^2}\sqrt{1-y^2}} \mathrm{d} x \mathrm{d}y \\
        &\quad + \iint_{[-1,1]^2} \left(\sum_{n=1}^{\infty}\frac 2 n \frac{T_n(x)\left(\sum_{i=1}^{\infty} a_i T_i(x)\right)}{\pi \sqrt{1-x^2}} \frac{T_n(y)\left(\sum_{i=1}^{\infty} a_i T_i(y)\right)}{\pi \sqrt{1-y^2}} \right) \mathrm{d}x \mathrm{d}y \\
        &= \log 2 \left(\int_{-1}^{1} \frac{1}{\pi \sqrt{1-x^2}} \mathrm{d} x\right)^2 \\
        &\quad + \sum_{n=1}^{\infty}\frac 2 n \left(\int_{-1}^{1} \frac{T_n(x)\left(\sum_{i=1}^{\infty} a_i T_i(x)\right)}{\pi \sqrt{1-x^2}} \mathrm{d}x \right) \left(\int_{-1}^{1} \frac{T_n(y)\left(\sum_{i=1}^{\infty} a_i T_i(y)\right)}{\pi \sqrt{1-y^2}} \mathrm{d}y \right)  \\
    \end{aligned}
    \]
    We recall the orthogonal relation of Chebyshev polynomials, 
    \[
    \int_{-1}^{1} \frac{T_a(x) T_b(x)}{\pi \sqrt{1-x^2}} \mathrm{d} x = 
    \left\{
    \begin{aligned}
        &1, \quad &a=b=0; \\
        &\frac{1}{2}, \quad &a=b\ne 0; \\
        &0, \quad &a\ne b.
    \end{aligned}
    \right.
    \]
    Using the orthogonal relation, we further simplify $F(\mu)$ as
    \[
    \begin{aligned}
        F(\mu) &= \log 2 \left(\int_{-1}^{1} \frac{1}{\pi \sqrt{1-x^2}} \mathrm{d} x\right)^2 \\
        & \quad + \sum_{n=1}^{\infty}\frac 2 n \left(\int_{-1}^{1} \frac{a_n T_n(x)T_n(x)}{\pi \sqrt{1-x^2}} \mathrm{d}x \right) \left(\int_{-1}^{1} \frac{a_n T_n(y)T_n(y)}{\pi \sqrt{1-y^2}} \mathrm{d}y \right)  \\
        &= \log 2 + \sum_{n=1}^{\infty} \frac{a_n^2}{2n}. \\
    \end{aligned}
    \]
    Therefore, the problem $\inf_\mu F(\mu)$ admits a simple solution: $a_i = 0$ for every $i \ge 1$,
    and 
    \[ \begin{aligned}
        \mathrm{d} \mu_0 &= \frac{\mathrm{d} x}{\pi \sqrt{1-x^2}}; \\
        F(\mu_0) &= \log 2.
    \end{aligned} \]

    Now we solve the equilibrium measure under one additional constraint $\int x\mathrm{d} \mu \ge r$: 
    \[\inf_{\mu:\ \int x\mathrm{d} \mu \ge r} F(\mu). \]
    Since
    \[
    \int_{-1}^{1} \mathrm{d} \mu = \frac{T_1(x)}{\pi \sqrt{1-x^2}} \sum_{i=0}^{\infty} a_i T_i(x) \mathrm{d} x = \frac{a_1}{2} = r,
    \]
    this suggests that the equilibrium measure $\mu_r$ is
    \[
    \mu_r (x) = \frac{1 + 2rx}{\pi \sqrt{1-x^2}} \quad (r \le 1/2),
    \]
    and 
    \[\inf_{\mu:\ \int x\mathrm{d} \mu \ge r} F(\mu) = F(\mu_r) = \log 2 + 2r^2. \]

    Now we are close to completion: our rate function $\tilde{I}(\mu)$ is computed as
    \[
    \begin{aligned}
        \tilde{I}(\mu) &= F(\mu) - c = F(\mu) - F(\mu_0) = F(\mu) - \log 2.
    \end{aligned}
    \]
    We conclude that the rate is
    \[ 
    \begin{aligned}
        &\exp \left(- m^2 \cdot \inf _{\mu:\ \int x\mathrm{d} \mu \ge r} \tilde{I} (\mu) \right) \\
        &= \exp \left(- m^2 \cdot (\inf _{\mu:\ \int x\mathrm{d} \mu \ge r} \tilde{F} (\mu) - F(\mu_0)) \right) \\
        &= \exp\left( - m^2 \cdot 2r^2 \right),
    \end{aligned}
    \]
    which shows a good rate function for $r$: $I(r) = 2r^2$.

    The function $\mu_r$ induces a measure if and only if $r \le 1/2$. 
    When $r > 1/2$, the function $\mu_r$ no longer induces a measure, as it violates positivity near $x=-1$. 
    This suggests that our theoretical bound, $\log 2 + (2r)^2$, cannot be attained here 
    (To satisfy the positivity of $\mu$, the coefficients $a_2, a_3, \cdots$ cannot all be zero simultaneously).

    Overall, we have
    \[ \inf _{\mu:\ \int x\mathrm{d} \mu \ge r} \tilde{I} (\mu) \ge (2r)^2\]
    and
    \[ 
    \left\{ \begin{aligned}
    I(r) &= 2r^2, \quad &\text{for } r \le \frac12; \\
    I(r) &> 2r^2, \quad &\text{for } r > \frac12. 
    \end{aligned} \right.\]

\end{proof}

\subsection{Additional Notes on the $\mathrm{O}(2m) \setminus \mathrm{SO}(2m)$ Case}
When $A \in \mathrm{O}(2m) \setminus \mathrm{SO}(2m)$, $\det(A) = -1$. There are two eigenvalues fixed at
$+1$ and $-1$, respectively.

Our formula becomes:
\[p(t_1, \cdots, t_{m-1})\mathrm{d}t_1 \cdots\mathrm{d}t_{m-1}  = C' \prod_{1\leq k<j\leq m-1}(t_k - t_j)^2 \cdot \prod_{1\leq i \leq m-1} (1-t_i^2)^{+1/2} \mathrm{d}t_1 \cdots\mathrm{d}t_{m-1}.\]
Taking the logarithm, we have
\[-\log p(t_1, \cdots, t_{m-1}) = \sum_{1\leq k<j\leq m-1} \left(-2 \log|t_k - t_j| \right) - \sum_{1\leq i \leq m-1} \left(\frac{\log (1-t_i^2)}{2}\right) + C_0.\]
The only differences from above are that $m$ is replaced by $m-1$ and the external field term is negated. 
Both changes are negligible in the thermodynamic limit $m \to \infty$. Thus, we obtain the same conclusion for the rate function.

\subsection{The Exact Form of Rate Function $I(r)$}
We conjecture that the exact form of $I(r)$ is:
\[ 
    \left\{ \begin{aligned}
    I(r) &= 2r^2, \quad &\text{for } r \in \left(0, \frac{1}{2}\right]; \\
    I(r) &= \frac{1}{2} - \log 2 - \log(1-r), \quad &\text{for } r \in \left[\frac{1}{2}, 1\right). 
    \end{aligned} \right.
\]
Like the Gross-Witten Ensemble \citep{GrossWitten1980}, there is a third-order phase transition near $r= 1/2$.

\section{Layer-wise Homology}
\label{sec_comparison}
We compare our method against REEF (Representation Encoding Fingerprint) in revealing layer-wise homology \citep{zhang2025reef} on the following two model pairs:
\begin{itemize}
    \item \textbf{Llama-3.1-8B vs. Llama-3.2-1B} \citep{meta2024llama32}: A pruned version where we expect a structured layer correspondence;
    \item \textbf{Llama-3.1-8B vs. Qwen3-8B-Base} \citep{yang2025qwen3technicalreport}: Two independently developed models with no known lineage.
\end{itemize}
\begin{figure}[htbp]
    \centering
    \subfigure[MDIR, Llama-3.1-8B vs. Llama-3.2-1B]{
        \includegraphics[width=0.565\textwidth]{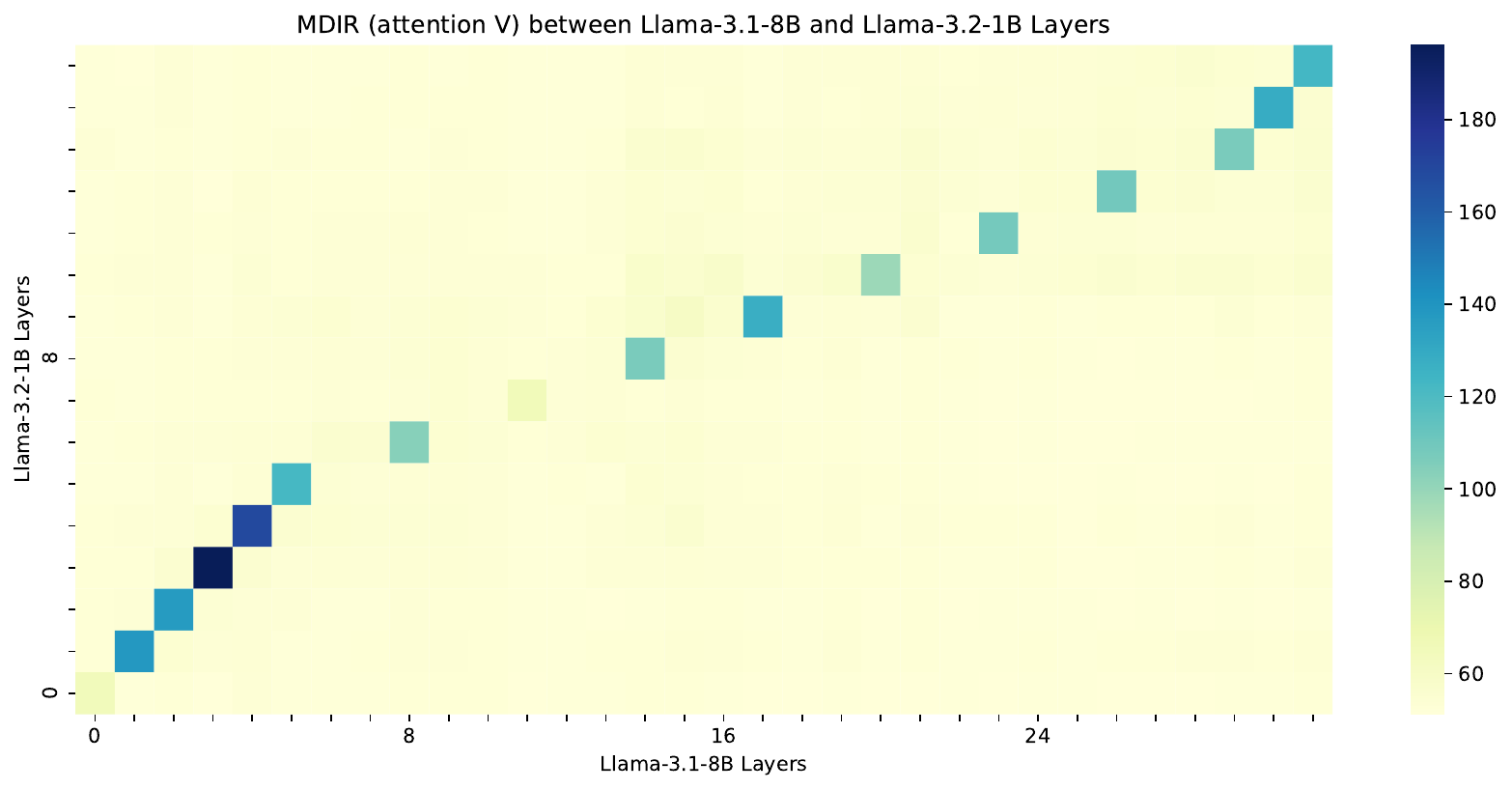}
    }
    \subfigure[MDIR, Llama-3.1-8B vs. Qwen3-8B-Base]{
        \includegraphics[width=0.323\textwidth]{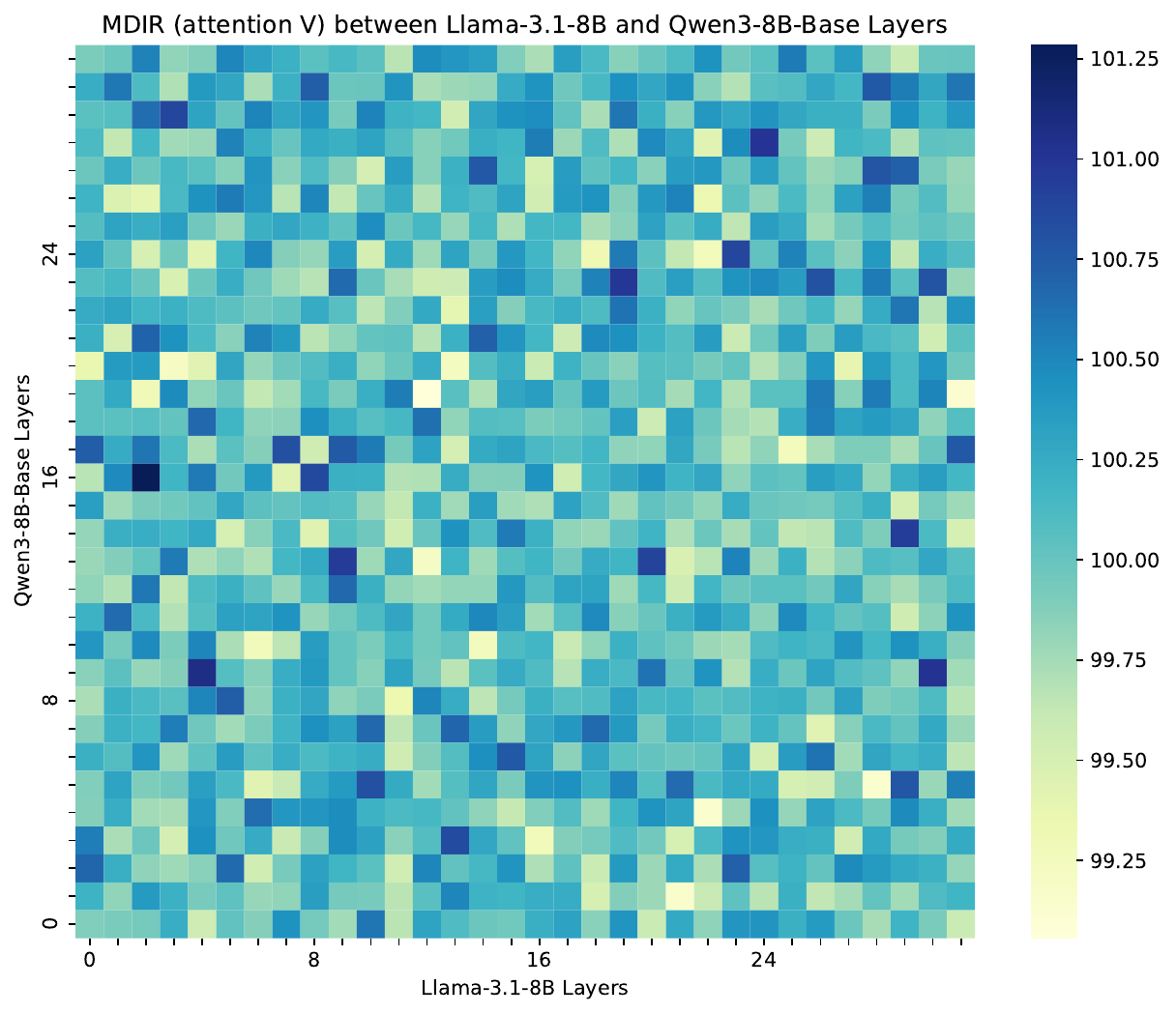}
    }
    \subfigure[REEF, Llama-3.1-8B vs. Llama-3.2-1B]{
        \includegraphics[width=0.565\textwidth]{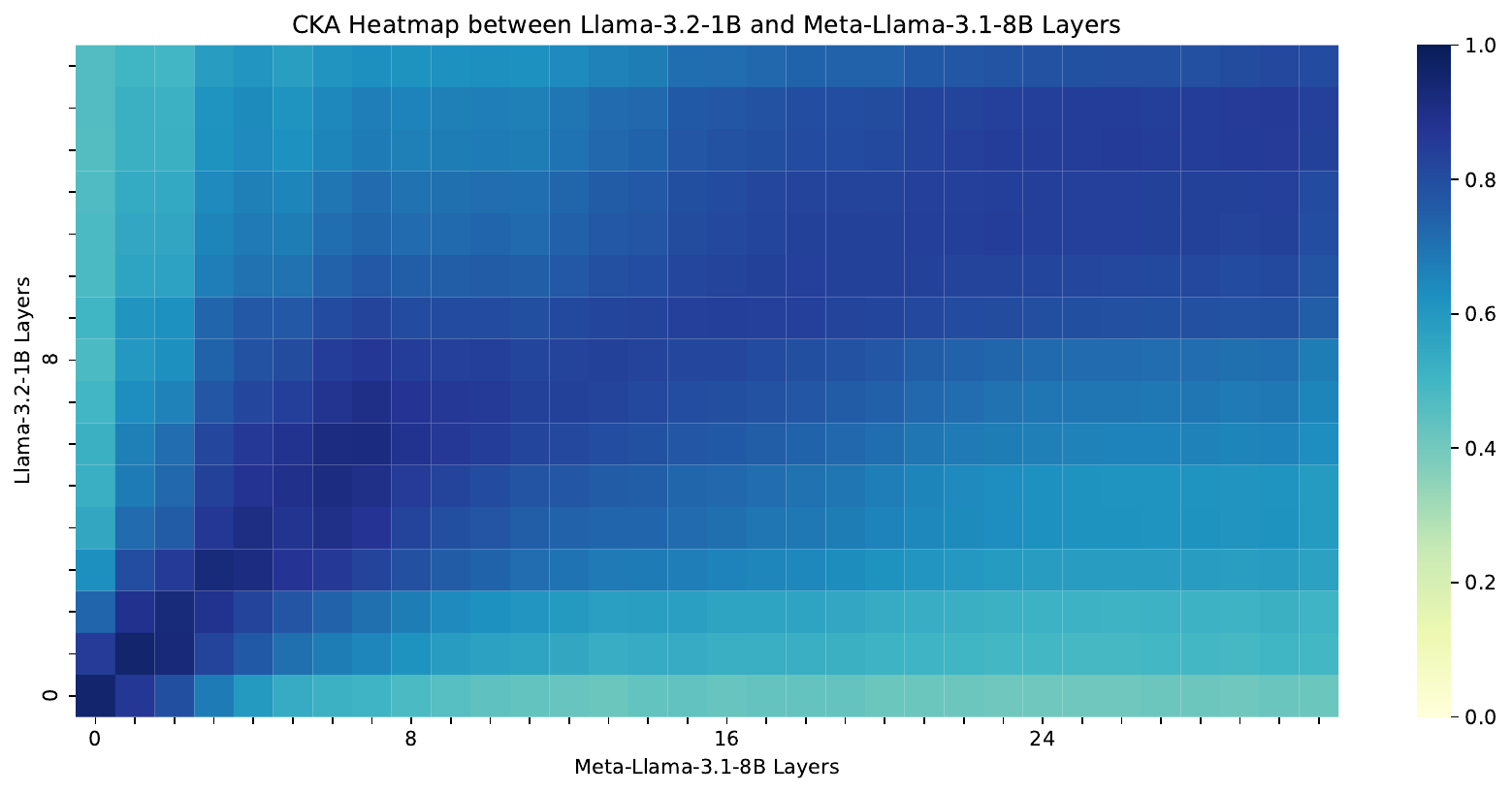}
    }
    \subfigure[REEF, Llama-3.1-8B vs. Qwen3-8B-Base]{
        \includegraphics[width=0.323\textwidth]{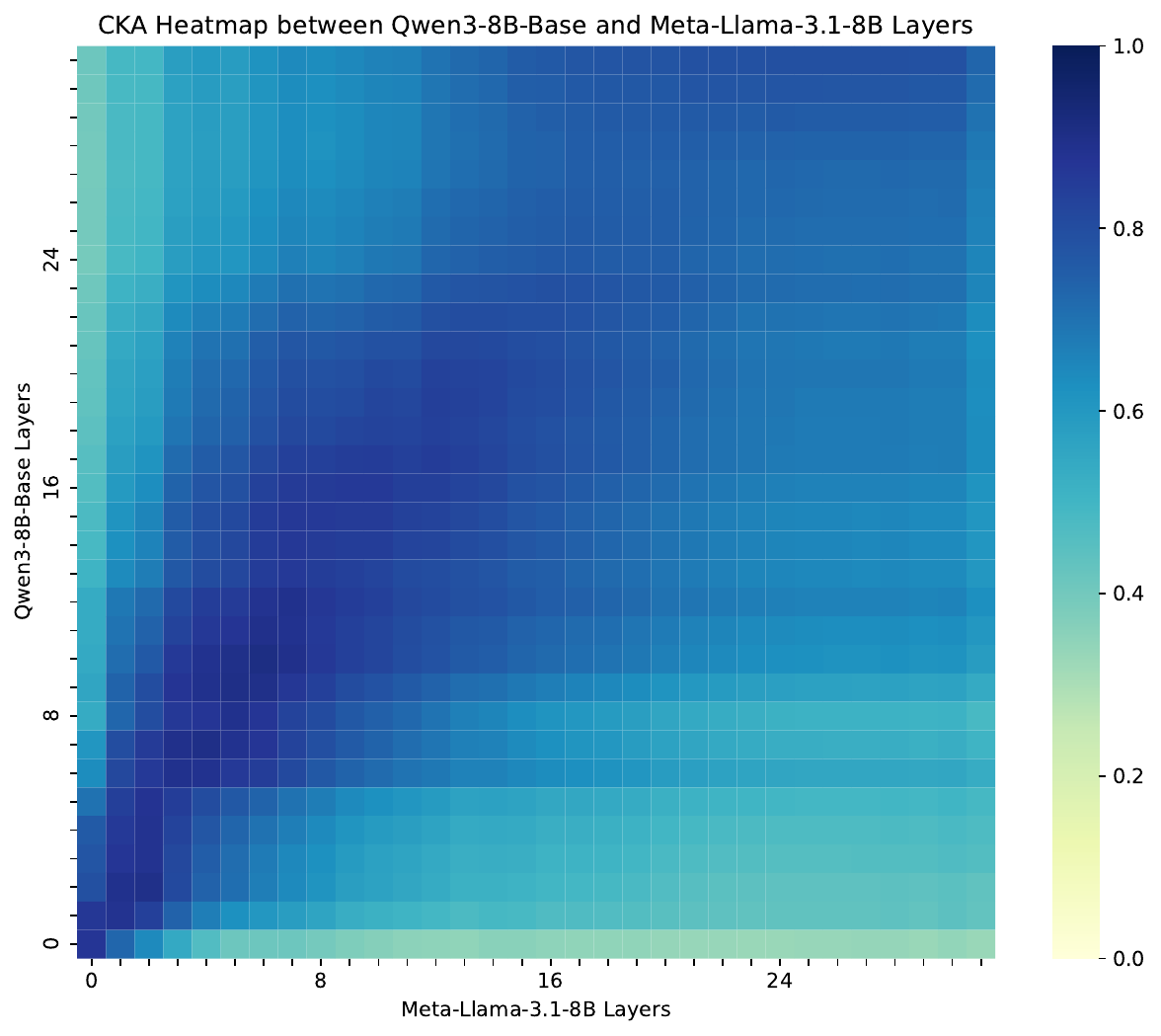}
    }
    \caption{MDIR vs. REEF on revealing layer-wise relasionship.}
    \label{fig:reef}
\end{figure}
In Figure~\ref{fig:reef}(a), MDIR produces a diagonal-like pattern, indicating that each layer of Llama-3.2-1B aligns with a specific layer in Llama-3.1-8B 
(suggesting the 16 layers of Llama-3.2-1B are derived from layer 0, 1, 2, 3, 4, 5, 8, 11, 14, 17, 20, 23, 26, 29, 30 and 31 of Llama-3.1-8B, respectively). 
This reveals not only homology but also the exact strategy used during model pruning. 
This offers MDIR a very fine-grained level of interpretability which previous methods cannot provide.
In contrast, REEF (Figure~\ref{fig:reef}(c)) yields uniformly high CKA similarity measures across many pairs of layers
and failing to identify the exact correspondence. 
While it captures global similarity, it lacks the granularity to reveal which layers are actually aligned, thus less effective for model homology detection.
When applied to unrelated models (Figure~\ref{fig:reef}(b,d)), 
MDIR correctly outputs a noise pattern, consistent with the absence of structural homology. 
REEF, however, continues to report high similarities ($>0.9$) between certain layer pairs.
Our results demonstrate that MDIR is capable of interpretably reconstructing layer mappings between related model weights.

\end{document}

%% file: math_commands.tex
\usepackage{amsmath,amsfonts,bm}

\def\eqref#1{equation~\ref{#1}}

\def\1{\bm{1}}

\DeclareMathAlphabet{\mathsfit}{\encodingdefault}{\sfdefault}{m}{sl}
\SetMathAlphabet{\mathsfit}{bold}{\encodingdefault}{\sfdefault}{bx}{n}

\newcommand{\R}{\mathbb{R}}

\DeclareMathOperator*{\argmax}{arg\,max}

\DeclareMathOperator{\Tr}{Tr}